\documentclass[letterpaper, 10 pt, conference]{ieeeconf}  

\usepackage[table]{xcolor}

\usepackage{multirow}
\usepackage{times}

\usepackage{hyperref}

\usepackage{color}
\usepackage{algorithm}
\usepackage{algpseudocode}
\usepackage{listings}
\usepackage{courier}
\usepackage{soul}
\usepackage{amsmath,amsfonts,amssymb}

\usepackage{mathtools}
\usepackage{subfigure}
\usepackage{wrapfig}
\usepackage{hyperref}
\usepackage{multirow}
\usepackage{graphicx}
\usepackage{graphicx}
\usepackage{soul}
\usepackage[flushleft]{threeparttable}
\usepackage{enumerate}
\usepackage{algorithmicx}
\usepackage{algpseudocode}
\usepackage{epstopdf}

\IEEEoverridecommandlockouts                              
\definecolor{red}{rgb}{1.00,0.00,0.00}
\definecolor{blue}{rgb}{0.00,0.00,1.00}
\definecolor{green}{rgb}{0.2,0.70,0.2}
\definecolor{yellow}{rgb}{0.5,0.5,0.0}
\definecolor{white}{rgb}{1,1,1}

\title{\LARGE \bf
A Hierarchical Framework to Generate Robust Biped Locomotion Based on Divergent Component of Motion
\author{Mohammadreza Kasaei, Nuno Lau and Artur Pereira
 \\IEETA / DETI University of Aveiro 3810-193 Aveiro, Portugal \\
	\{mohammadreza, nunolau, artur\}@ua.pt
}
}

\begin{document}
\maketitle
\thispagestyle{empty}
\pagestyle{empty}
\begin{abstract}	
Keeping the stability can be counted as the essential ability of a humanoid robot to step out of the laboratory to work in our real environment. Since humanoid robots have similar kinematic to a human, humans expect these robots to be robustly capable of stabilizing even in a challenging situation like while a severe push is applied. This paper presents a robust walking framework which not only takes into account the traditional push recovery approaches (e.g., ankle, hip and step strategies) but also uses the concept of Divergent Component of the Motion~(DCM) to adjust next step timing and location. The control core of the proposed framework is composed of a Linear-Quadratic-Gaussian~(LQG) controller and two proportional controllers. In this framework, the LQG controller tries to track the reference trajectories and the proportional controllers are designed to adjust the next step timing and location that allow the robot to recover from a severe push. The robustness and the performance of the proposed framework has been validated by performing a set of simulations, including walking and push recovery using~\mbox{MATLAB}. The simulation results verified that the proposed framework is capable of providing a robust walking even in very challenging situations.
\end{abstract}
\section{Introduction}
\label{sec:introduction}
Unsafeness of humanoid robots is the main reason for preventing these robots from stepping out of the laboratory and adapting to our environment. Since some decades ago, many types of research have been conducted and the capability of humanoid robots has been much improved to perform stable walking but it does not still satisfy human expectations. The question is \textit{how is it that a human is adept at performing walking even in very challenging conditions but humanoid robots are only capable of performing slow walking?}

Generating a fast and stable walking for humanoid robots is a multidisciplinary and complex subject due to the naturally unstable dynamics of these types of robots. To reduce the complexity of this subject, it is decomposed into several small independent modules generally. Then these modules will be connected hierarchically to generate a walking system. A common decomposition approach is using a simplified physical model of the robot's dynamics and developing a walking system based on this model. In this approach, the overall structure of the walking system is decoupled into four hierarchy levels which are footstep planner, reference generators, push recovery and low-level controller. Using this structure has gained great attention because of reducing the complexity, increasing the flexibility and also the portability of the walking system~\cite{kasaei2017reliable}.

Linear Inverted Pendulum Model~(LIPM)~\cite{kajita2003biped} is one of the successful simplified physical models among the others which is capable of generating a feasible reference trajectory of the Center Of Mass~(COM). This model restricts the overall dynamics into COM and it is able to generate a fast, efficient and feasible trajectory for the COM movement according to a set of pre-planned footsteps. Using this model, the desired trajectory of COM can be generated based on one of the analytical solution, preview control or Differential Dynamic Program~(DDP), which are appropriate for real-time implementation~\cite{kajita2003biped}. After generating the reference trajectories, the major task of the low-level controller is tracking the reference trajectories and compensating the tracking error to keep the stability of the robot. Indeed, the controller tries to control either the Zero Moment Point~(ZMP) or Divergent Component of Motion~(DCM)~\cite{englsberger2015three}. The low-level controller has been formulated successfully using classical feedback controllers, Linear Quadratic Regulator~(LQR)-based methods and also Model Predictive Control~(MPC). ZMP-based controllers try to keep the robot's stability by keeping the ZMP inside the support polygon. DCM-based approaches split the LIPM dynamics into stable and unstable parts and just by controlling the unstable part, keep the stability of the robot~\cite{takenaka2009real}. Although both methods can handle small and normal tracking errors, in case of large disturbance, ZMP-based methods are not able to handle the large tracking error due to the limited size of the support polygon. DCM-based methods try to handle such situations by changing the landing location of the swing leg.

In this paper, we tackle the problem of designing a robust walking framework based on DCM which takes into account adjusting the next step timing and location. The fundamental component of the proposed framework is a Linear Quadratic Gaussian~(LQG) which not only can optimally track the reference trajectory in the presence of noise and disturbances but also can illuminate the steady-state error. Furthermore, by measuring the DCM error at each control cycle, two proportional controllers are designed to adjust the landing time and location of the swing leg to increase the withstanding level of the robot. The remainder of this paper is structured as follows: Section~\ref{sec:related_works} gives an overview of related work. In Section~\ref{sec:arct}, the overall architecture of the proposed system is presented and the functionality of each module is explained and verified. In Section~\ref{sec:simulation}, two simulation scenarios are designed to verify the performance of the proposed system. Based on the simulation results, discussion and comparison will be given in Section~\ref{sec:DISCUSSIONS}. Finally, conclusions and future research are presented in Section~\ref{sec:CONCLUSION}.

\section {Related Work}
\label{sec:related_works}
Many types of research have been conducted to realize the full potential of biped robots and they showed a robust balance recovery is an essential part of a biped walking system. DCM-based approaches are one of the successful and appropriate methods to develop a robust walking controller that can react against unpredictable external pushes. These approaches decouple the dynamics of COM into divergent and convergent components and, just by controlling the divergent component try to keep the stability of the robot~\cite{takenaka2009real,englsberger2015three,englsberger2017smooth,kamioka2018simultaneous}. Some of these researches will be reviewed briefly in the remainder of this section.

Pratt et al.~\cite{pratt2006capture} proposed an extended version of LIPM, which is composed of a flywheel to consider the momentum around the COM. According to this dynamics model, they defined the Capture Point~(CP) concept that is a point on the ground where the robot should step to keep its stability. Using their model, Stephens~\cite{stephens2007humanoid} determined a decision surface to responses to external disturbances. The decision surface breaks the recovery reactions into three particular strategies which were ankle, push and step strategy. Based on this decision surface, in case of small disturbances, ankle strategy is used to compensate for the error using shifting the Center of Pressure~(COP) to apply more ankle torques. However, the COP is limited by the size of support polygon, the ankle strategy is not useful in the case of larger disturbances. In the case of a large disturbance, robot should use the hip and waist joints to generate angular momentum around the COM for stabilizing the states of the system~(hip strategy). Similar to the ankle strategy, the maximum feasible angular momentum is limited and in the case of very large disturbances, robot should take steps to keep its stability. 

Morisawa et al.~\cite{morisawa2014biped} proposed online walking generators with push recovery by utilizing the CP concept which was based on a PID controller. The performance of their method has been verified using performing a set of simulations with the~\mbox{HRP-2} humanoid robot. Simulation results showed that the robot could perform walking on the uneven terrain while keeping its stability.

Hopkins et al.~\cite{hopkins2014humanoid} released the height constraint of LIPM and considered a dynamics model based on time-varying DCM and showed how a generic COM height trajectories could be generated by modifying the natural frequency of the DCM during stepping. They designed a walking on an uneven terrain scenario in a simulation environment to validate the performance of their method. The simulation results verified the capability of their method.

Englsberger et al.~\cite{englsberger2015three} proposed the 3D version of CP and introduced the Enhanced Centroidal Moment Pivot point~(eCMP) and the Virtual Repellent Point~(VRP). They showed how eCMP and VRP could encode the magnitude, direction of the external impact. They designed a closed-loop motion tracking control verified the robustness of the controller with regarding the several uncertainties using performing a set of simulations and also real experiments.

Kryczka et al.~\cite{kryczka2015online} proposed an online biped locomotion planner based on a nonlinear optimization technique that is capable of modifying the next step position and timing to keep the stability during walking. Their approach has been validated using performing some real experiments on the real humanoid platform COMAN. The results showed that online modifying the step position and timing could increase the recovery level of the robot in the face with external disturbances.

Khadiv et al.~\cite{6930e2c2955c4eb688fa0507e3bc2032} combined step location and timing adjustment together to generate robust gaits. Their approach is composed of two main stages. The first stage is responsible for specifying the nominal step location and step duration at the beginning of each step. In the second stage, the landing location and time of the swing leg will be modified at each control cycle according to the measured DCM using an optimization method. The performance of their method has been verified using different simulation scenarios. The simulation results showed that step time adjustment improved the robustness.

Griffin et al.~\cite{griffin2016model} designed an MPC to generate a stable dynamics walking based on a time-varying DCM. They used step positions and rotations as the control inputs and considered some reachability constraints to ensure that the generated step positions are kinematically reachable. They demonstrated the performance of their method by performing fast and stable walking with footstep adjustment using the simulated ESCHER humanoid robot. Shafiee-Ashtiani et al.~\cite{shafiee2017robust} also showed a robust walking could be formulated as an MPC based on time-varying DCM concept.
\begin{figure*}[t]
	\begin{centering}
		\begin{tabular}	{c}	
			\hspace{15mm}\includegraphics[width=0.85\linewidth,trim= 0cm 9cm 0cm 0cm,clip] {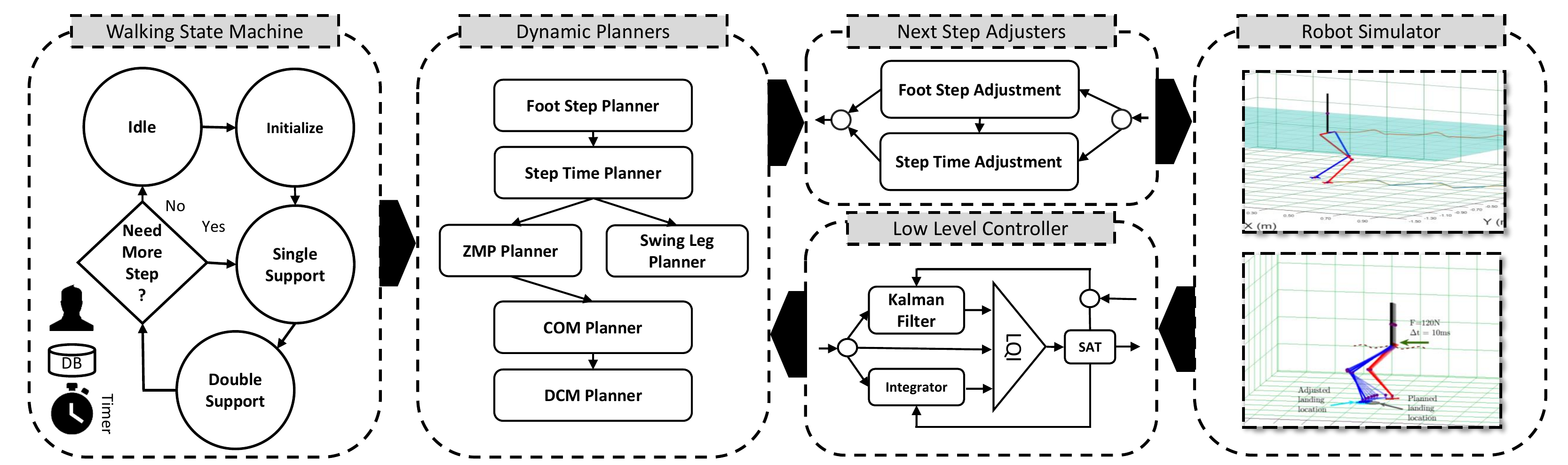}		
		\end{tabular}	
	\end{centering}			
	\caption{ The overall architecture of the proposed framework.  }
	\vspace{-2mm}
	\label{fig:overall}
\end{figure*}

Kamioka et al.~\cite{kamioka2017dynamic} proposed a dynamic gait re-planing method by applying the cyclic gait criterion based on the DCM. Their method does not only take footsteps positions and its timing into account but also it considers the locomotion mode~(e.g., walking, running, and hopping). They used an approximated gradient descent algorithm to modify of footsteps and timing. Their method has been validated using real robot experiments. Later, in~\cite{kamioka2018simultaneous}, the authors proposed a new quadratic programming~(QP) approach based on an analytical solution of DCM. By performing several real experiments, they showed that their method is capable of compensating disturbances in a hierarchical strategy.

Jeong et al.~\cite{jeong2017biped} introduced a closed-loop foot placement controller based on CP, which is capable of generating the desired ZMP according to the current CP error. They try to stabilize walking using developing disturbance adapting walking pattern generator based on CP, an ankle torque reference generator and also by adjusting the next footstep position and timing. Their method has been validated using performing a set of simulations in the Choreonoid simulator with a model of DRC-HUBO+.

\section{Architecture}
\label{sec:arct}
In this section, we summarize the components of our walking system presented in~Fig.~\ref{fig:overall}. Particularly, the proposed system is composed of four main modules which are explained in the remainder of this section. 
\subsection{Walking State Machine}
Due to the periodic nature of walking, it can be modeled by a state machine that state transitions are occurred based on an associated timer and also the conditions of the states. This state machine controls the overall process of the framework and composes of four main states which are depicted in Fig.~\ref{fig:overall}. In the \textit{Idle state}, the robot stops and waiting for a start walking command. Once a walking command received, the state transits to the \textit{Initialize state} and the robot tries to shift its COM to the first support foot to be ready for taking the first step. In \textit{Single Support State} and \textit{Double Support State}, walking trajectories are generated. It should be mentioned that the associated timer will be reset at the end of the double support state.

\subsection{Dynamic Planners}
This module is responsible for generating all the walking reference trajectories. This process is started by planning a set of footprints according to some constraints (e.g., maximum step size, the distance between feet) and the input command which can be either walking velocity or given step info parameters. After planning the footprints, the step time planner, plans the time of each step according to the input velocity and also the step time adjuster module. This planner determines the total time of each step, including the time of single and double support phases. According to the generated footprints and step times, the trajectories of ZMP and swing leg will be determined by the ZMP planner and the swing leg planner, respectively. The ZMP planning procedure is determined using the following formulation:
\begin{equation}
p= 
\begin{cases}
f_{i}  \qquad\qquad\qquad\qquad\qquad 0 \leq t < T_{ss} \\
f_{i}+ \frac{SL \times (t-T_{ss})}{T_{ds}} \qquad\qquad T_{ss} \leq t < T_{ss}+T_{ds} 
\end{cases} ,
\label{eq:zmpEquation}
\end{equation}
\noindent
where $p = [p_x \quad p_y]^\top$ is the generated ZMP, $t, T_{ss}, T_{ds}$ represent the time, duration of single and double support phases, respectively. $SL$ is the length of the step, $f_{i} = [f_{i,x} \quad f_{i,y}] \quad i \in \mathbb{N}.$ represents the planned foot prints on a 2D surface. As is mentioned before, $t$ will be reset at the end of each step~\mbox{$(t \geq T_{ss}+T_{ds}$)}.

In order to generate the swing trajectory, a Bezier curve is used to move the swing leg smoothly during lifting and landing based on a set of input parameters which are the maximum swing height parameter~($z_s$), the generated foot prints and the step times.

In our target framework, LIPM is used as our template dynamics model for generating the trajectory of COM and DCM. In the remainder of this section, it will be briefly reviewed and we will explain how the analytical solution of LIPM will be used to generate the reference trajectories of COM and DCM. 

LIPM considers some assumptions like restricting the vertical motion of COM to simplify and approximate the dynamics model of a humanoid robot using a first-order stable dynamics model as follows:
\begin{equation}
\ddot{c} = \omega^2 ( c - p) \quad ,
\label{eq:lipm}
\end{equation}
\noindent
where $c = [x_c \quad y_c \quad  z_c]^\top$ is the position of COM, \mbox{$\omega = \sqrt{\frac{g+\ddot{z}_c}{z_c}}$} represents the natural frequency of the pendulum, $p$ is the ZMP position. As is explained before, the reference trajectory of ZMP is already generated and it can be used to determine the position of the COM at the start and end of each step. Accordingly, by considering these positions as the boundary conditions for the (\ref{eq:lipm}), it can be solved analytically as a boundary value problem and the trajectory of COM is determined as follows:
\begin{equation}
\label{eq:com_traj_x0xf}
\resizebox{0.95\linewidth}{!} {$
	c(t) = p + \frac{ (p-c_f) \sinh\bigl((t - t_0)\omega\bigl)+ (c_0 - p) \sinh\bigl((t - t_f)\omega\bigl)}{\sinh((t_0 - t_f)\omega)}$ },
\end{equation}
\noindent
where $t_0$ and $t_f$ represent the starting and the ending time of a step, $c_0$, $c_f$ are the corresponding positions of COM at these times, respectively. After determining the COM trajectory, the DCM reference trajectory should be generated. Conceptually DCM defines a point that the robot should step to rest over the support foot~\cite{pratt2006capture} and it's dynamics is defined as follows:
\begin{equation}
\zeta = c + \frac{\dot{c}}{\omega} \quad ,
\label{eq:dcm}
\end{equation}
\noindent
where $\zeta = [\zeta_x \quad \zeta_y]^\top$ represents the 2D DCM,  $\dot{c}$ is the velocity of COM and $\omega$ is the natural frequency of the pendulum.

\begin{figure}[!t]
	\label {exPlanner}
	\begin{centering}
		\begin{tabular}	{c c}			
			\includegraphics[width = 0.465\columnwidth]{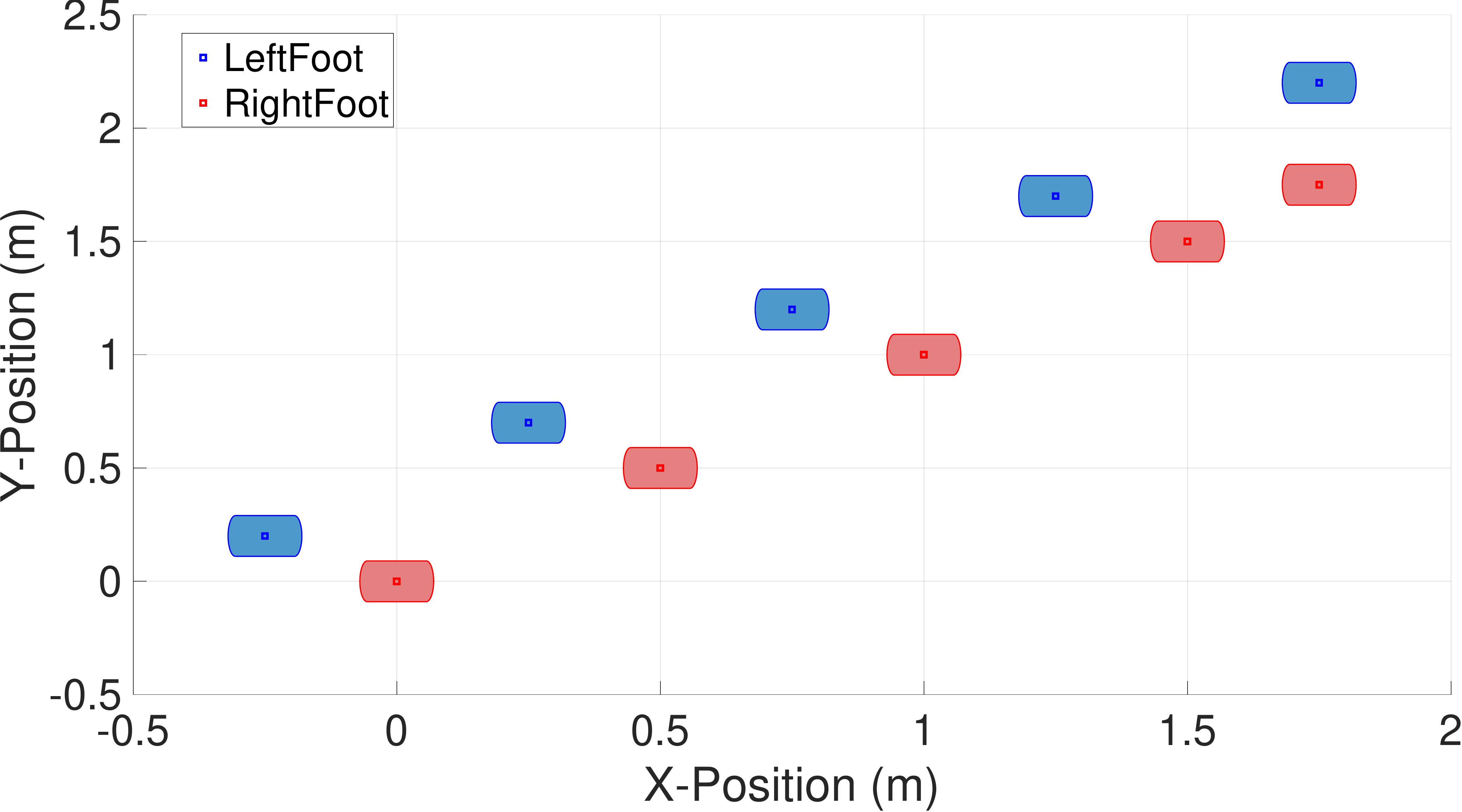}&
			\hspace{-2mm}\includegraphics[width = 0.465\columnwidth]{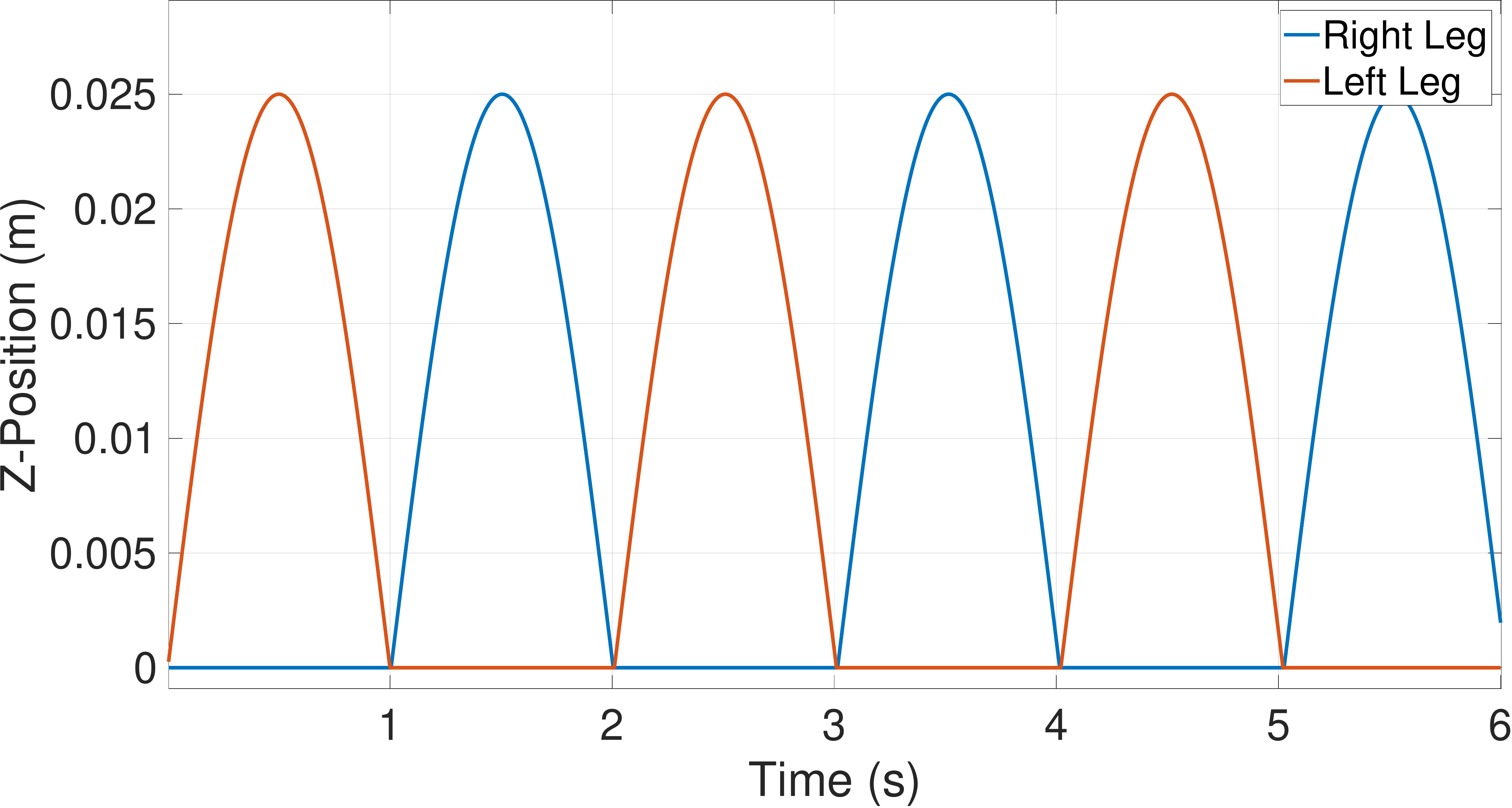}\\
			\includegraphics[width = 0.465\columnwidth]{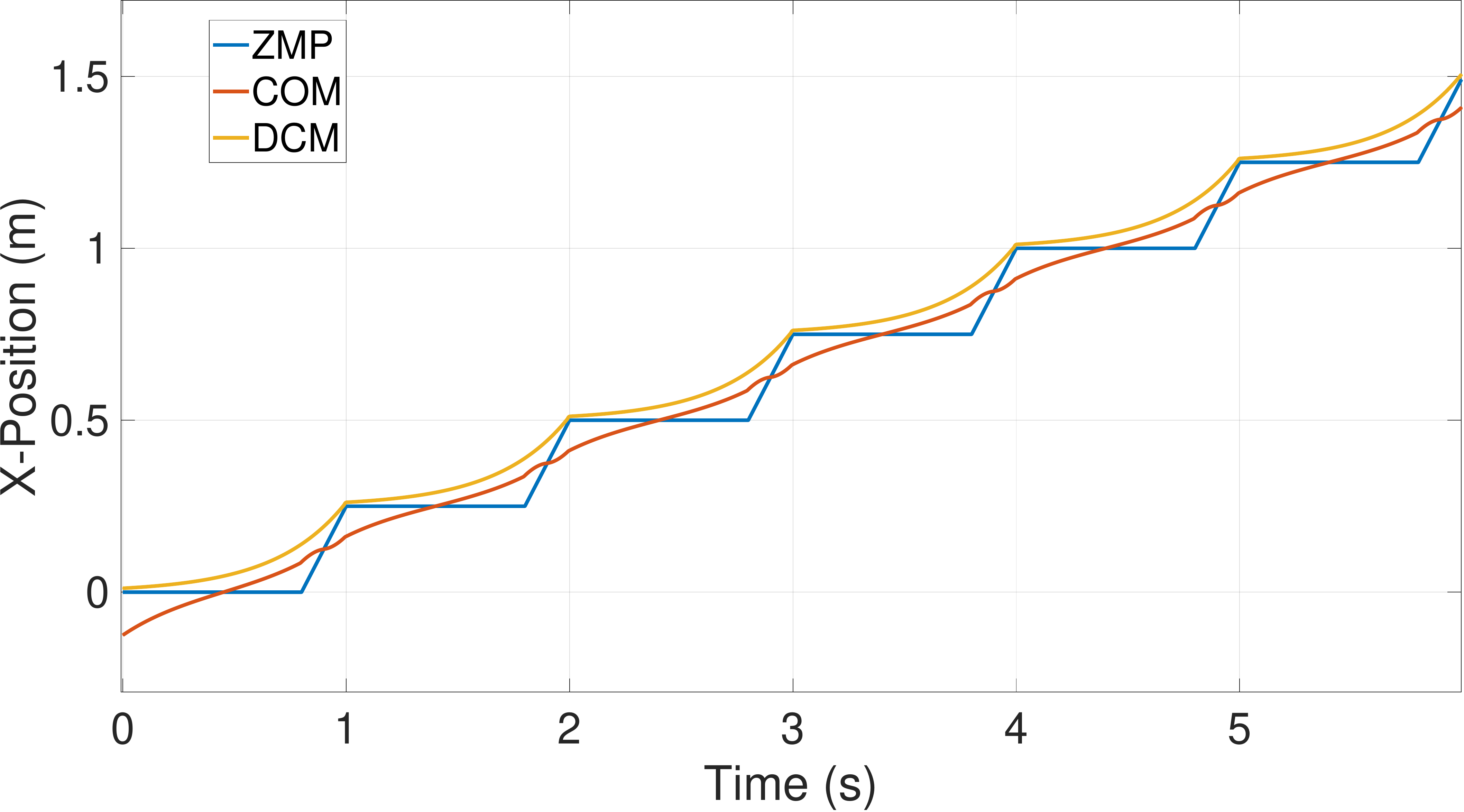}&
			\hspace{-2mm}\includegraphics[width = 0.465\columnwidth]{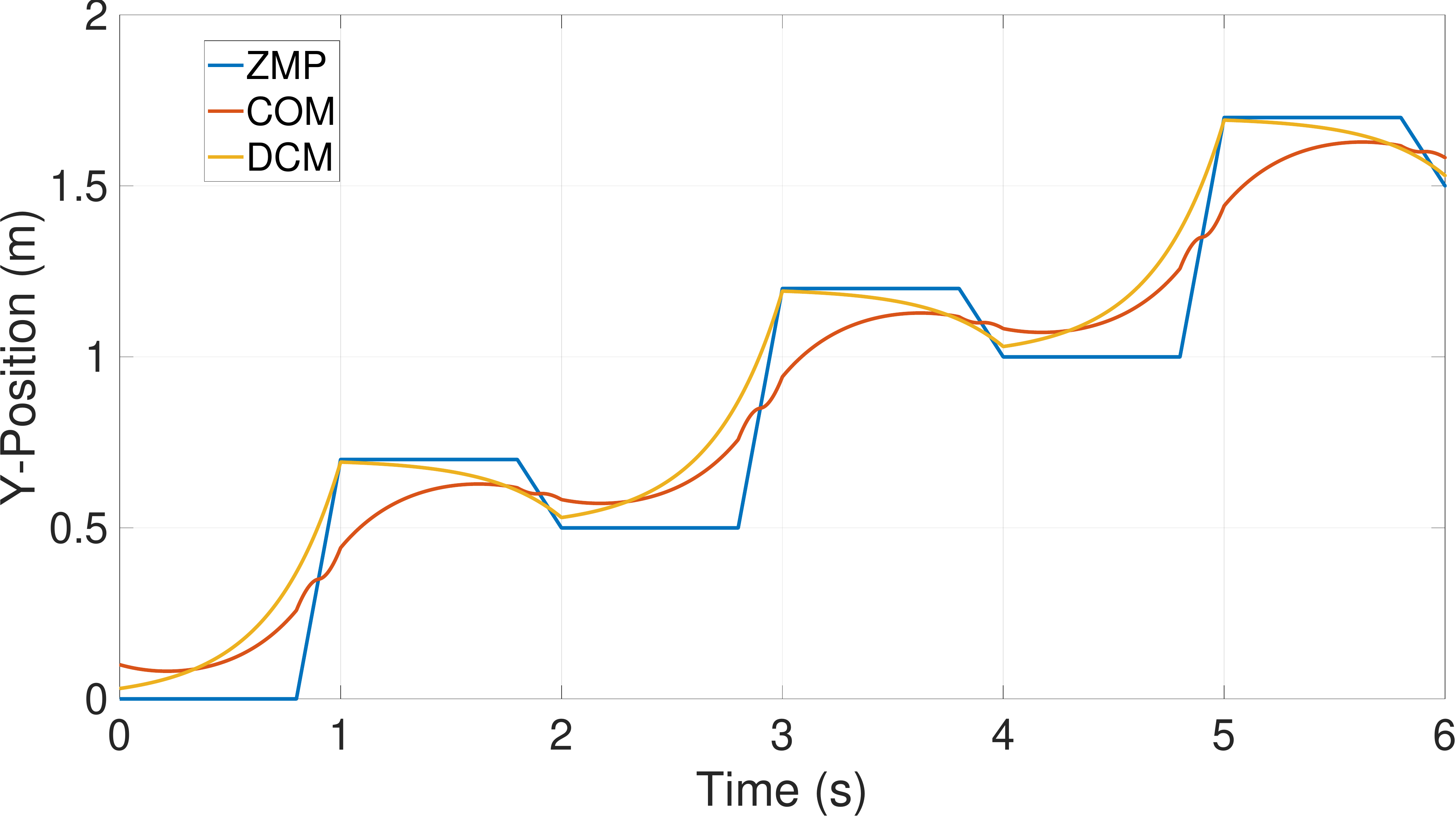}
		\end{tabular}
	\end{centering}
	\vspace{-0mm}
	\caption{ Planned reference trajectories for the exemplary diagonal six-steps walking. \textbf{\mbox{Top-left}}: planned foot prints, \textbf{\mbox{Top-right}}: generated swing leg trajectories, \textbf{\mbox{Bottom-left}}: corresponding ZMP, COM and DCM reference trajectories in X, \textbf{\mbox{Bottom-right}}: corresponding ZMP, COM and DCM reference trajectories in Y.}
	\vspace{-5mm}
	\label{fig:exPlanner}
\end{figure}

To validate the performance of the planners in this module, an exemplary 6-steps walking has been planned. In this example, a simulated robot should walk diagonally with a step length~($SL$) of 0.5$m$~(equal in both X and Y directions), duration of single support~($T_{ss}$) and double support~($T_{ds}$) are considered to be 0.8$s$ and 0.2$s$, respectively. The maximum swing leg height~($z_s$) is assumed to be 0.025$m$ and the height of COM~($z_c$) is 1$m$. The results of the walking scenario are shown in Fig.~\ref{fig:exPlanner}. As results showed, this module is capable of generating the walking reference trajectories according to the input parameters.

\subsection{Low Level Controller}
\begin{figure}[!t]
	\label {excontrol}
	\begin{centering}
		\begin{tabular}	{c c}			
			\includegraphics[width = 0.465\columnwidth]{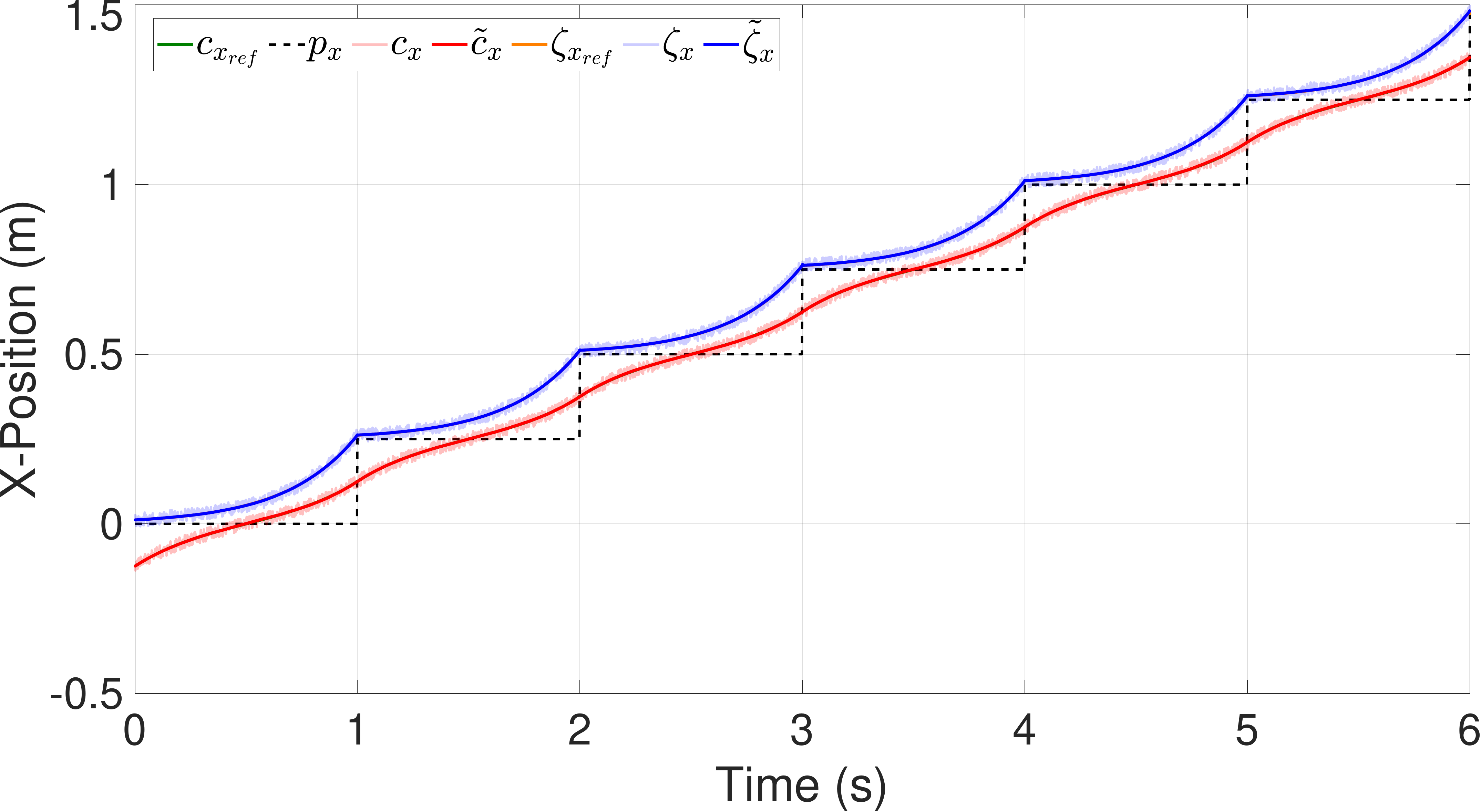}&
			\hspace{-2mm}\includegraphics[width = 0.465\columnwidth]{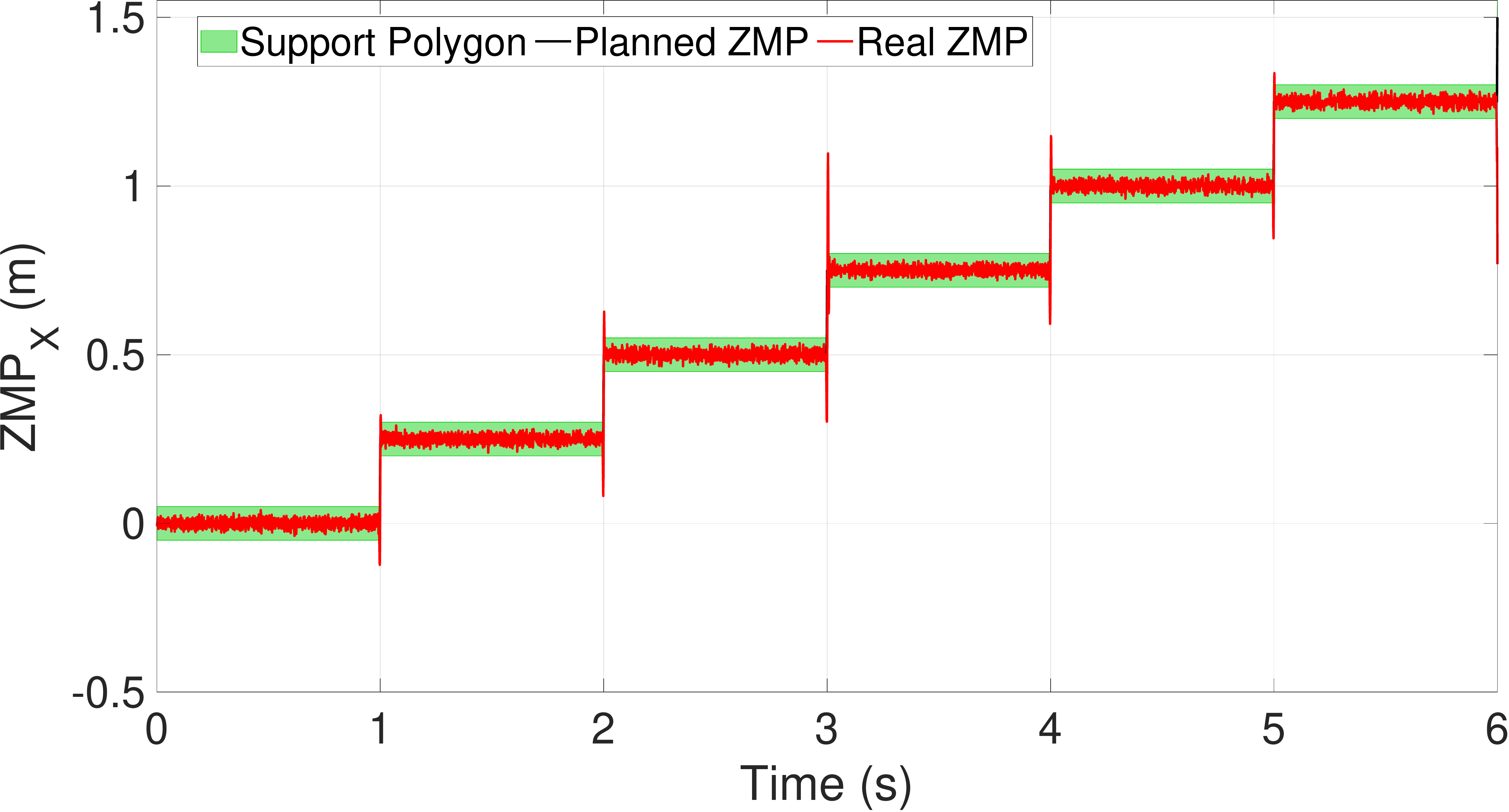}\\
			\includegraphics[width = 0.465\columnwidth]{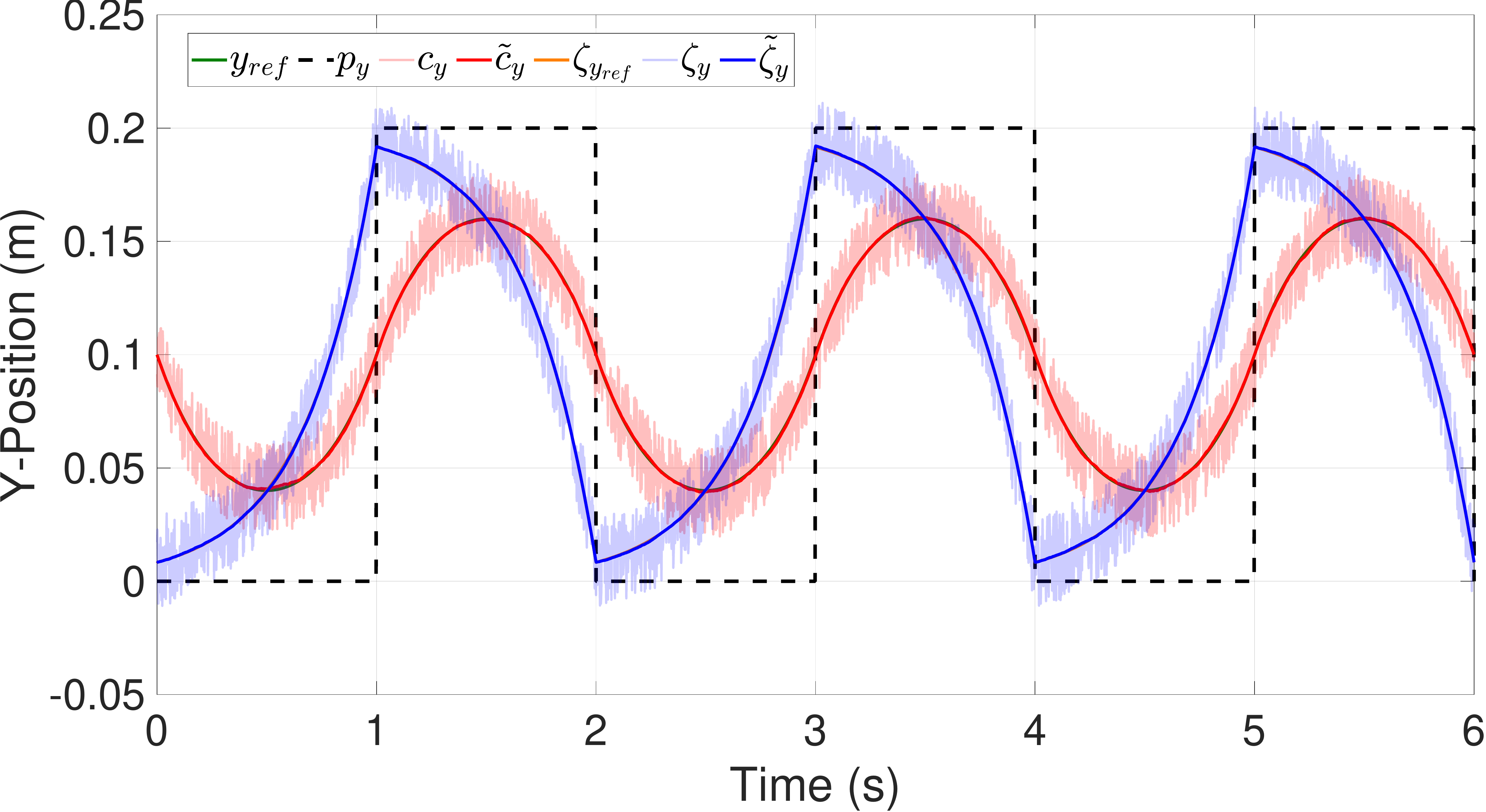}&
			\hspace{-2mm}\includegraphics[width = 0.465\columnwidth]{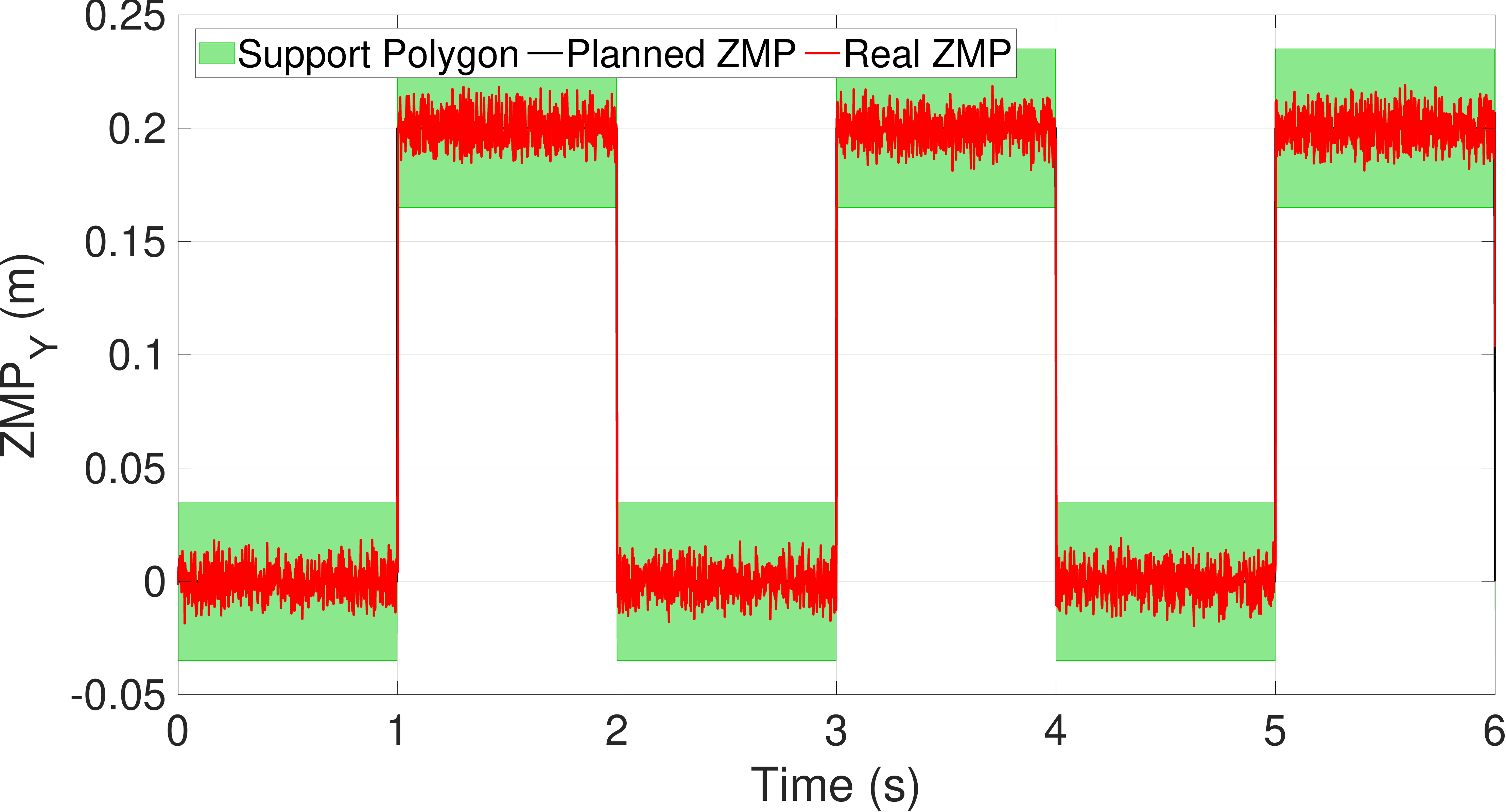}
		\end{tabular}
	\end{centering}
	\vspace{-0mm}
	\caption{ Simulation results of examining performance of the proposed controller in presence of measurement noise, $\mathcal{N}(\text{0, 6.25e-4})$. \textbf{\mbox{Top-left}}: reference and actual trajectories in X direction, \textbf{\mbox{Top-right}}:  reference and actual ZMP trajectories in X direction , \textbf{\mbox{Bottom-left}}: reference and actual trajectories in Y direction, \textbf{\mbox{Bottom-right}}: reference and actual ZMP trajectories in Y direction.}
	\vspace{-5mm}
	\label{fig:excontrol}
\end{figure}

The task of this module is to keep tracking the generated references even in the presence of noise and uncertainties. In order to develop this controller, LIPM and DCM are used to design a state-space system. According to the~(\ref{eq:dcm}), by taking derivative from both sides of this equation and then by substituting~(\ref{eq:lipm}) into the obtained result, a linear dynamics system will be obtained which can be represented in a state-space system as follows:
\begin{equation}
\frac{d}{dt} \begin{bmatrix} c \\ \zeta \end{bmatrix}
= 
\begin{bmatrix} 
-\omega & \omega \\ 
0 & \omega \\ 
\end{bmatrix}	
\begin{bmatrix} c \\ \zeta \end{bmatrix}
+
\begin{bmatrix} 
0 \\
-\omega
\end{bmatrix} p \quad ,
\label{eq:statespace_zeta}
\end{equation}
\noindent
as this system shows, the COM does not need to be controlled and it is always converged to the DCM. Hence, just by controlling the DCM, the system can always be stable. In this dynamics system, COM and DCM are the states of the system and they are considered to be measurable at each control cycle. Based on this system, a Linear-Quadratic-Gaussian~(LQG) controller is designed which is able to track the references robustly. As shown in the low-level controller module of Fig.~\ref{fig:overall}, the controller is composed of a Kalman Filter~(KF) to estimate the state of the system and eliminate the effect of measurement noise. Moreover, using the measurement of the system's state, an integrator is used to cancel the steady-state error. The fundamental component of this controller is an optimal state-feedback gain which is designed as follows:
\begin{equation}
u = -K
\begin{bmatrix}
\tilde{X} - X_{des}\\
X_i
\end{bmatrix} \quad ,
\end{equation}
\noindent
where $\tilde{X}$ is the output of the KF, $X_{des}, X_i$ represent the desired trajectories and the output of the integrator, respectively. $K$ is a gain matrix which defines the optimal action according to the measured and the desired state. Indeed, this gain should be designed such that the controller is capable of tracking the reference with minimizing the following cost function:
\begin{equation}
J(u) = \int_{0}^{\infty} \{ \phi^\intercal Q \phi + u^\intercal R u \} dt \quad ,
\end{equation}
\noindent
where $\phi = [\tilde{X} \quad X_i]^\intercal $, $Q$ and $R$ are selected based on trial and error commonly in order to balance the tracking performance and cost of control effort. It should be noted that a direct solution exists to find the $K$ matrix according to the $Q$ and $R$. To verified the performance and the robustness of the proposed controller, a simulation has been carried out. In this simulation, a six-step forward walking trajectories is generated ($SL_x$ = 0.5$m$, $SL_y$ = 0.0$m$, $T_ss = 1$, $T_ds =0$) and the controller should track the reference trajectories in presence of measurement noise. To examine the performance of the system, the state of the system and also real ZMP have been recorded while the simulated robot is walking (sampling rate = 500 Hz). The simulation results are depicted in Fig.~\ref{fig:excontrol}. The results showed that the controller could track the references even in the presence of noise.

\subsection{Next Step Adjusters}
In some situations like when a strong push is applied to the robot, the ZMP goes outside the support polygon and the low-level controller can not regain it back just by applying the compensating torques at the ankle and at the hip due to the saturation of these torques. Human combines two main strategies to cope with these situations: \textit{changing the landing location of swing leg} and \textit{adjusting the step time}. 

In our structure, these strategies have been developed in the \textit{Next Step Adjusters} module. To implement these strategies, the current measurement of DCM is used as an initial condition for (\ref{eq:dcm}) and this equation can be solved as an initial value problem to predict the landing position of the swing leg ($f_p$). The difference between the prediction and the next planned foot position is used as an error variable for the controllers of this module and it is defined as follows:
\begin{equation}
 \Delta f =  \underbrace{(\zeta_t - f_i)e^{w(T_{ss}-t)}}_{f_p}- f_{i+1} \quad .
\label{eq:dcm_delta_f}
\end{equation}
A proportional controller is designed based on $\Delta f$ in order to adjust the next footstep location:
\begin{equation}
 \delta p  =  -k_{sa}\Delta f \quad,
\label{eq:controller}
\end{equation}
\noindent
where $k_{sa}$ is the controller gain, $\delta p$ is the next step adjustment output which should be added to the output of the footstep planner in \textit{Dynamic Planners} module. It should be mentioned that a compliance margin is defined to avoid unnecessary adjustments. Moreover, the output of the footstep planner is saturated in order to prevent planning a footstep outside the kinematically reachable area for the robot. Due to the speed limitation of the swing leg, this strategy is useful if the robot has enough time for reaching to the new landing location. In some cases, like while the robot is in a constrained environment or when it does not have enough time to change the landing location, the step time adjustment can be used as a recovery strategy. In such situations, human either decreases the step time to rapidly put its foot down or increases it to regain its stability. This strategy can be combined with a step adjusting strategy to improve recovery performance. According to the (\ref{eq:dcm_delta_f}), step time and DCM are exponentially related together. Therefore, if we consider $T_{ss} = T_{ss}+\Delta t$, the step time adjustment ($\Delta t$) can be calculated using the following equation:
\begin{equation}
\Delta t  = \frac{1}{\omega} \log_e (\frac{\Delta f - f_{i+1}}{\zeta_t - f_i}) + t-T_{ss}  \quad,
\label{eq:steptime}
\end{equation}
\noindent
it should be noted that a first-order lag filter is used to avoid supper quick change in the step time adjustment because it can cause discontinuities in the generated swing leg position:
\begin{equation}
T_{ss} = T_{ss}(1-k_f) + (T_{ss}+\Delta t)k_f\quad,
\label{eq:lagfilter}
\end{equation}
\noindent
where $k_f$ specifies the effect of the step time changing in the current step time and it is determined by trials and error generally.

\begin{figure}[!t]
	\label {dcmscheme}
	\begin{centering}
		\begin{tabular}	{c}			
				\includegraphics[width = 0.95\columnwidth,trim= 0cm 8.5cm 11cm 0.45cm,clip]{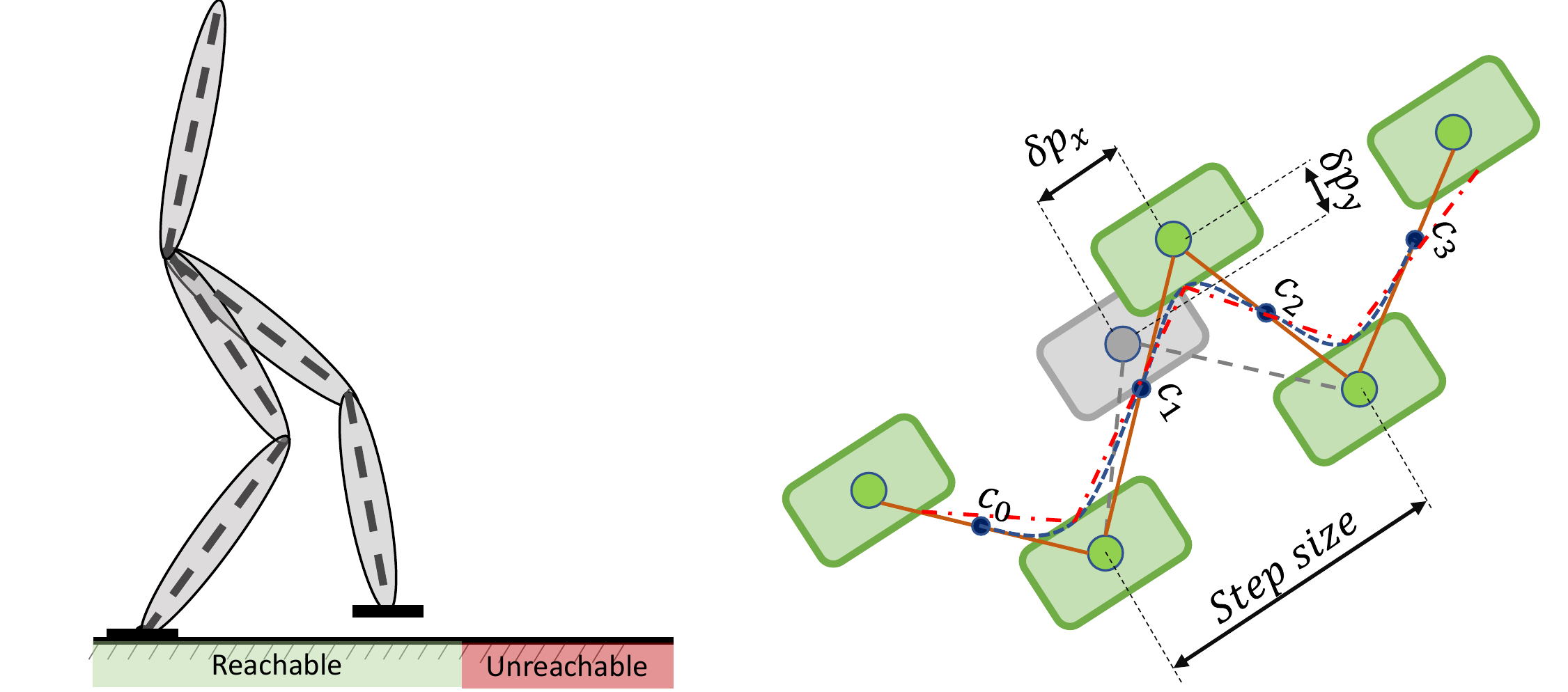}			
		\end{tabular}
	\end{centering}
	\vspace{-0mm}
	\caption{ Schematic of adjusting the step. \textbf{Left}: Kinematically reachable and unreachable regions for modifying the landing location of the swing leg. \textbf{Right:} an exemplary situation that the step adjuster modifies the landing location of the swing leg according to the measurement DCM.}
	\vspace{-5mm}
	\label{fig:dcmscheme}
\end{figure}

\section{Simulation}
\label{sec:simulation}
In this section, a set of simulation scenarios has been designed and carried out to validate the performance of the proposed framework. The simulations have been performed using a simulated robot which is developed in \mbox{MATLAB}. The physical property of the simulated robot is summarized in the following table:

\begin{table}[h!]
	\centering
	\caption{ Physical parameters of the simulated robot.}
	\vspace{-2mm}
	\begin{tabular}{|c|c|c|c|} 
		\hline
		mass & height of COM & foot length & foot width\\
		\hline
		30$kg$ & 1$m$  &0.15$m$ & 0.075$m$ \\ 	
		\hline
	\end{tabular}
	\label{tb:pushParams}
\end{table}

It should be noted that since the equations in sagittal and frontal planes are equivalent, the simulation results will be shown just in the sagittal plane.

\subsection{Scenario1: Keeping the stability while standing in single foot:} 
\begin{figure}[!t]
	\label {push_hip_ankle}
	\begin{centering}
		\begin{tabular}	{c c}			
			\includegraphics[width = 0.45\columnwidth]{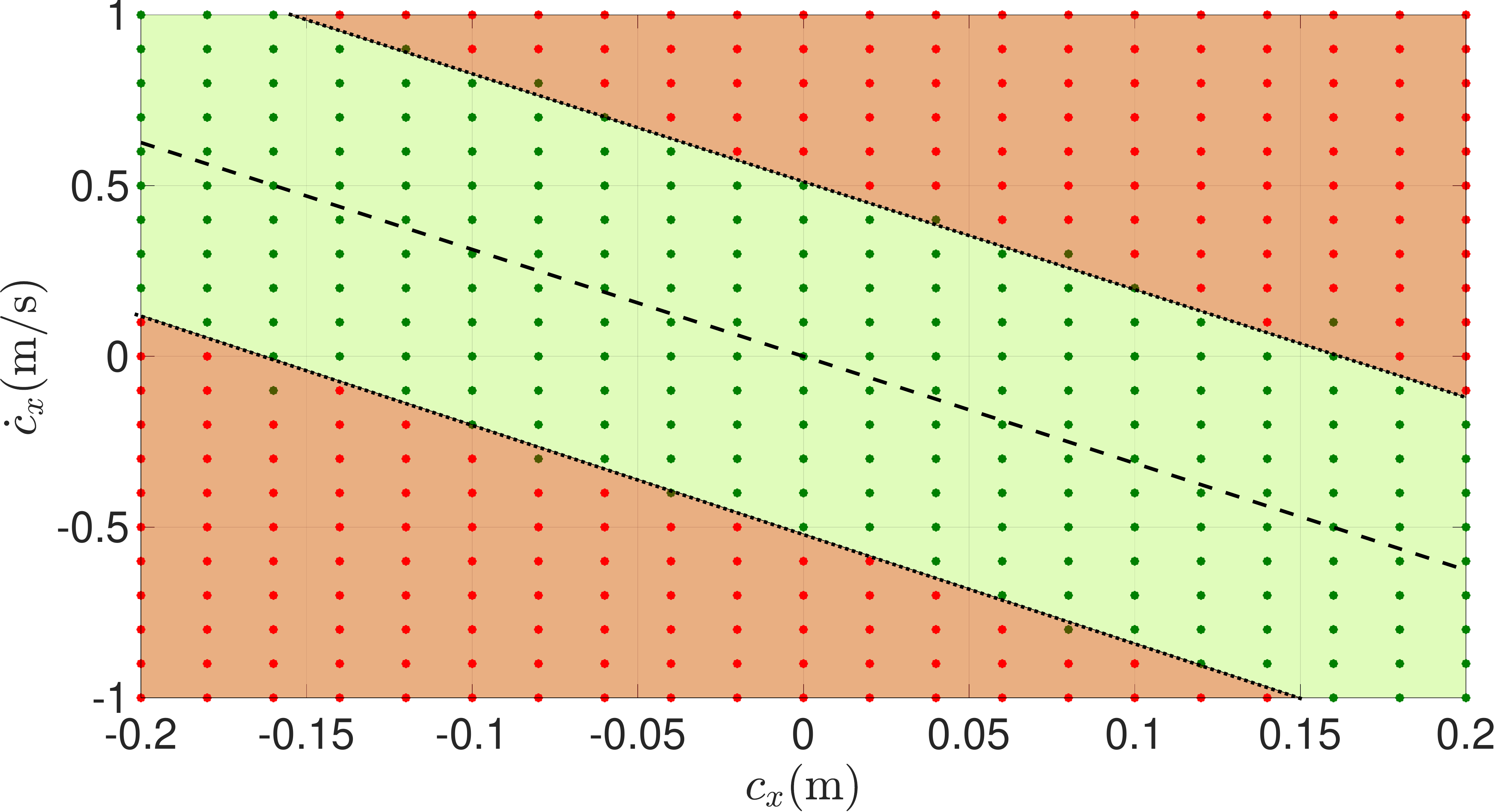} &
			\includegraphics[width = 0.45\columnwidth]{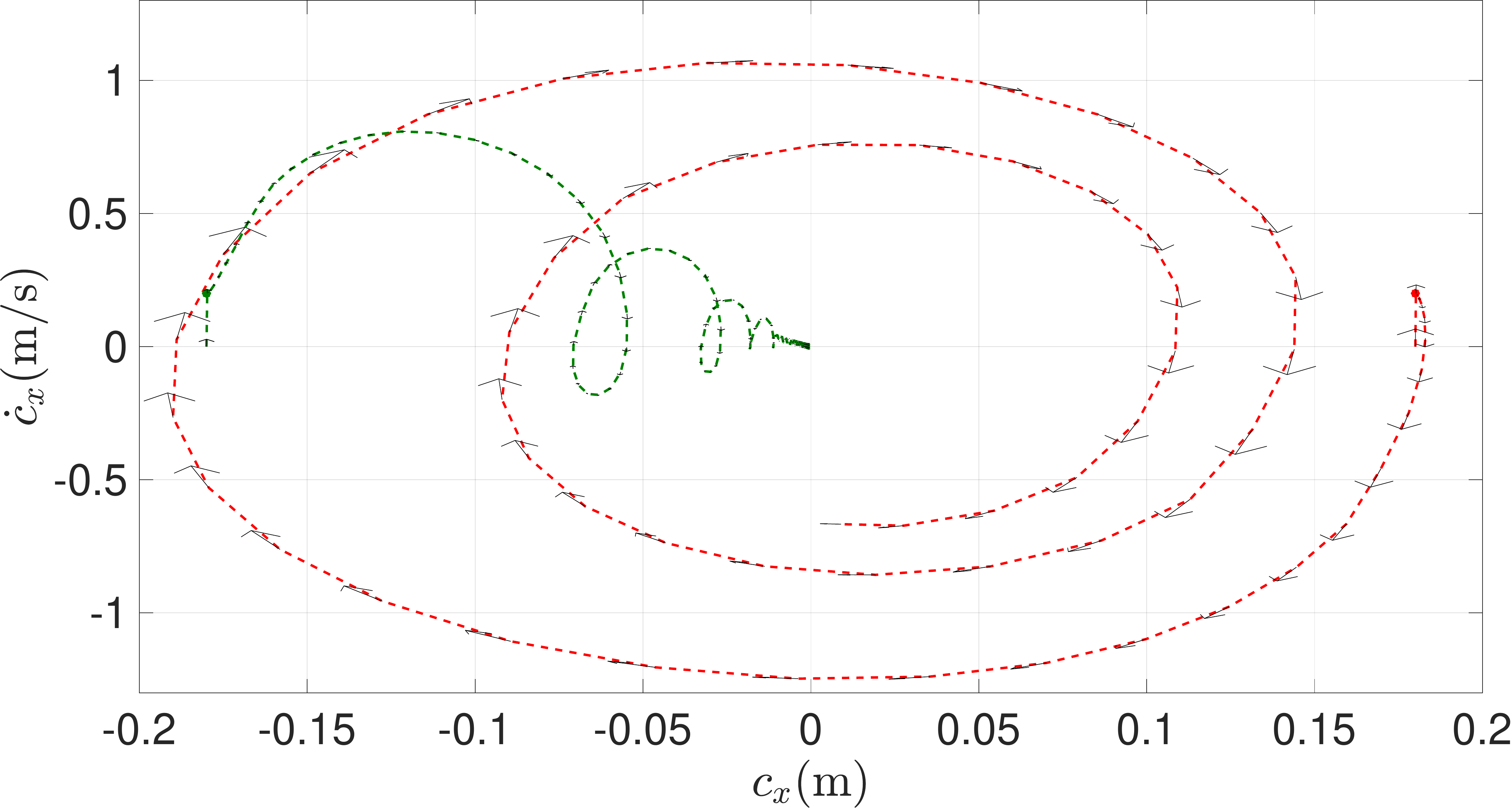} \\
			(a)&(b)			
		\end{tabular}
	\end{centering}
	\vspace{-2mm}
	\caption{ The simulations result of push recovery scenario. \textbf{(a)} Overall representation of the simulation results: \textbf{green region} represents the states that robot is able to keep its stability. \textbf{orange regions} represent the states that robot could not regain the stability. \textbf{dash-line} is $c+\frac{\dot{c}}{\omega} = 0$ which represents the most stable states, \textbf{(b)} two examples of the simulation results: \textbf{green-curve} represents a successful example, \textbf{red-curve} is an unsuccessful example. }  
	\vspace{-5mm}
	\label{fig:push_hip_ankle}
\end{figure}

The goal of this scenario is examining the performance of the low-level controller in regaining the stability of the robot during the single support phase. In this scenario, the simulated robot is considered to be stand in the single foot with a specified initial condition ($c_{x_0} $, $\dot{c}_{x_0}$) and the controller should control the states to return into ($0$,$0$) in maximum two seconds. Each simulation is started by selecting an initial condition over the range of [-0.2 0.2] at interval 0.02 $m$ for the position ($c_{x_0}$) and [-1 1] at interval 0.1 $m/s$ for the velocity ($\dot{c}_{x_0}$). According to these ranges, 441 simulations were performed and the results are graphically depicted in Fig.~\ref{fig:push_hip_ankle}. The left plot of this figure represents an overall result of the simulations. As is shown in this plot, if the initial state is selected from the green region, the controller can keep the stability; otherwise, the robot falls down. The right plots represent a successful simulation~(green curve) and an unsuccessful simulation~(red curve).

\subsection{Scenario2: Keeping the stability while walking in place:}
\begin{figure*}[!t]
	\centering
	\begin{tabular}	{c c c c}			
		\includegraphics[width=0.45\columnwidth, trim= 0cm 0cm 0cm 0cm,clip]{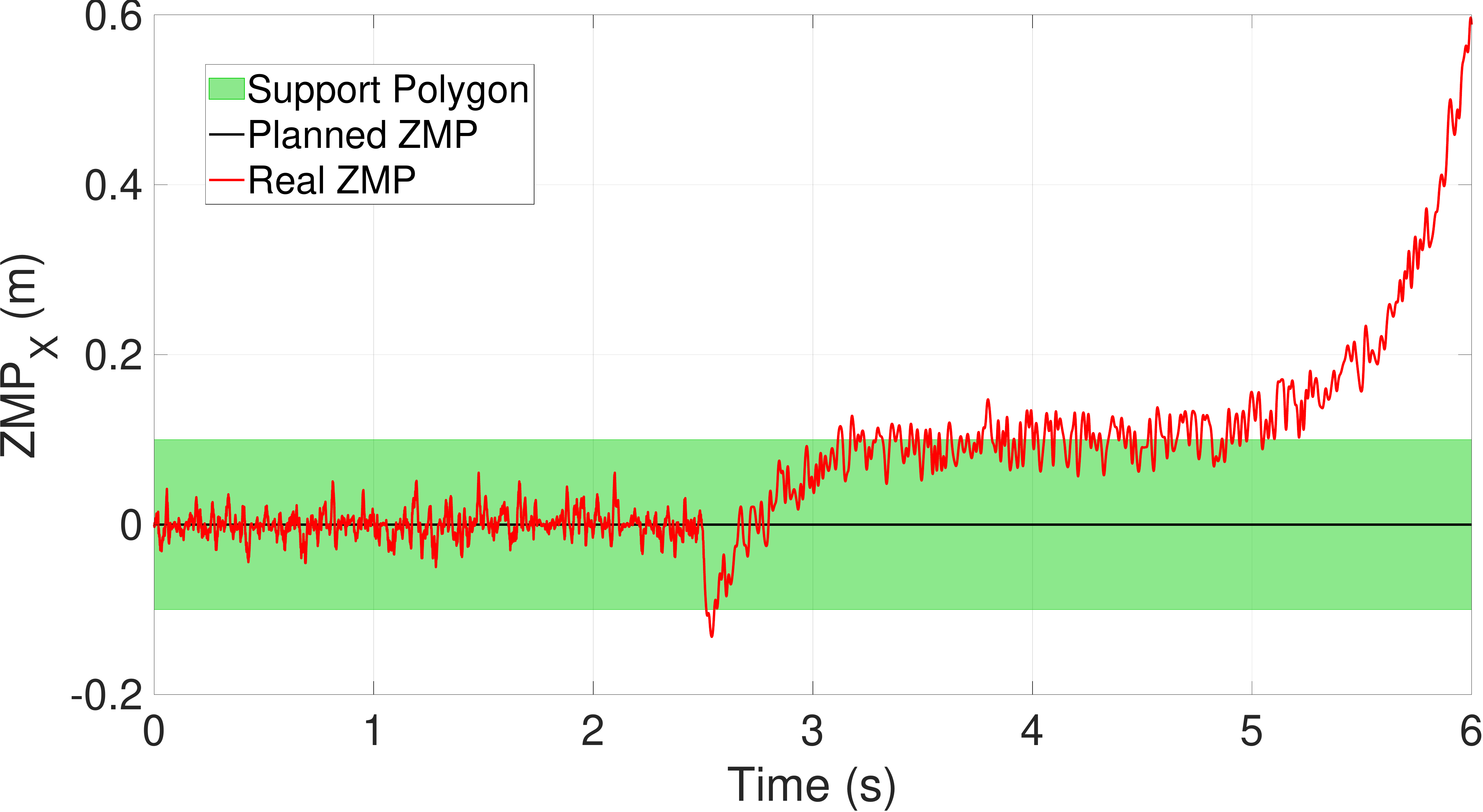}&
		\includegraphics[width=0.45\columnwidth, trim= 0cm 0cm 0cm 0cm,clip]{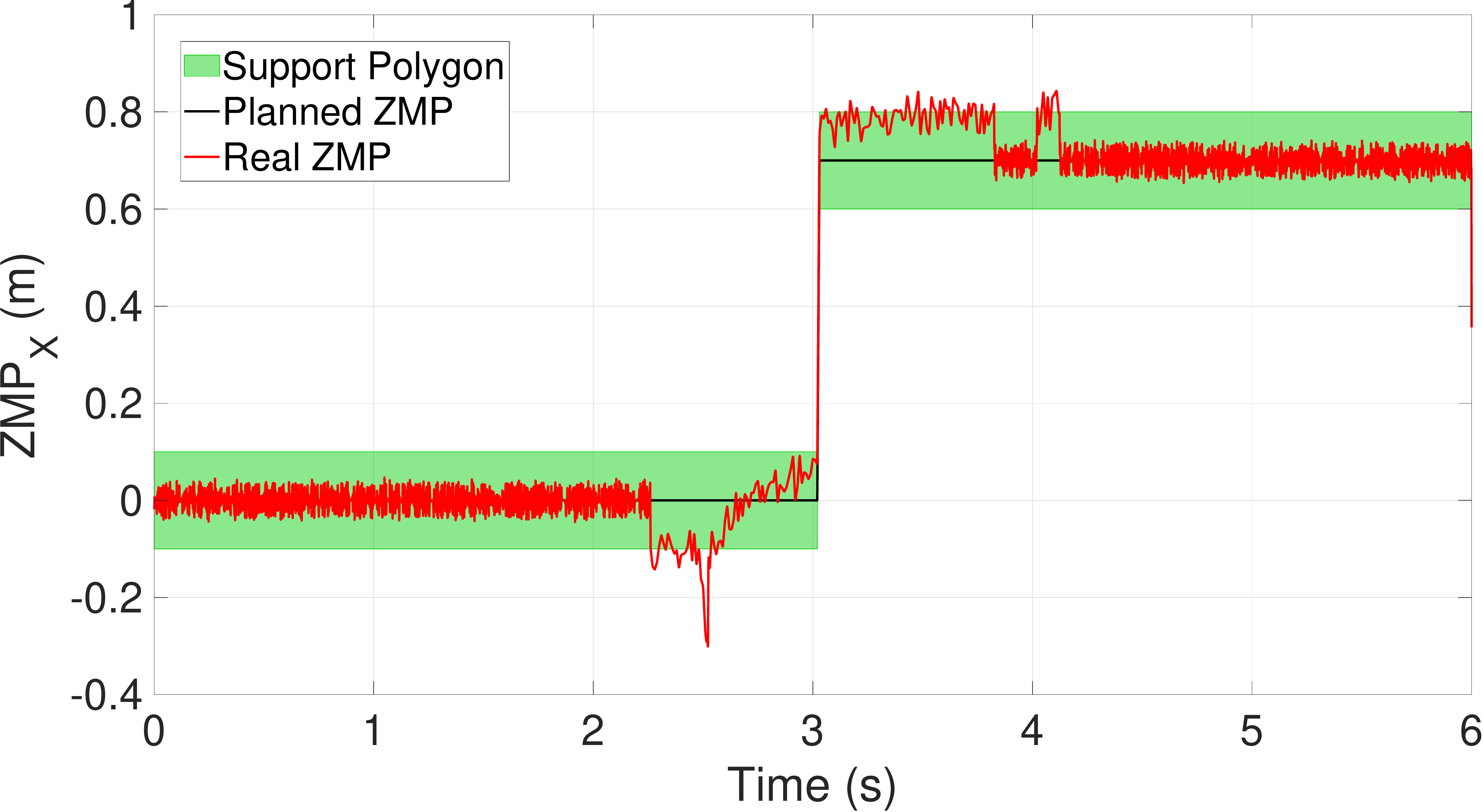}&
		\includegraphics[width=0.45\columnwidth, trim= 0cm 0cm 0cm 0cm,clip]{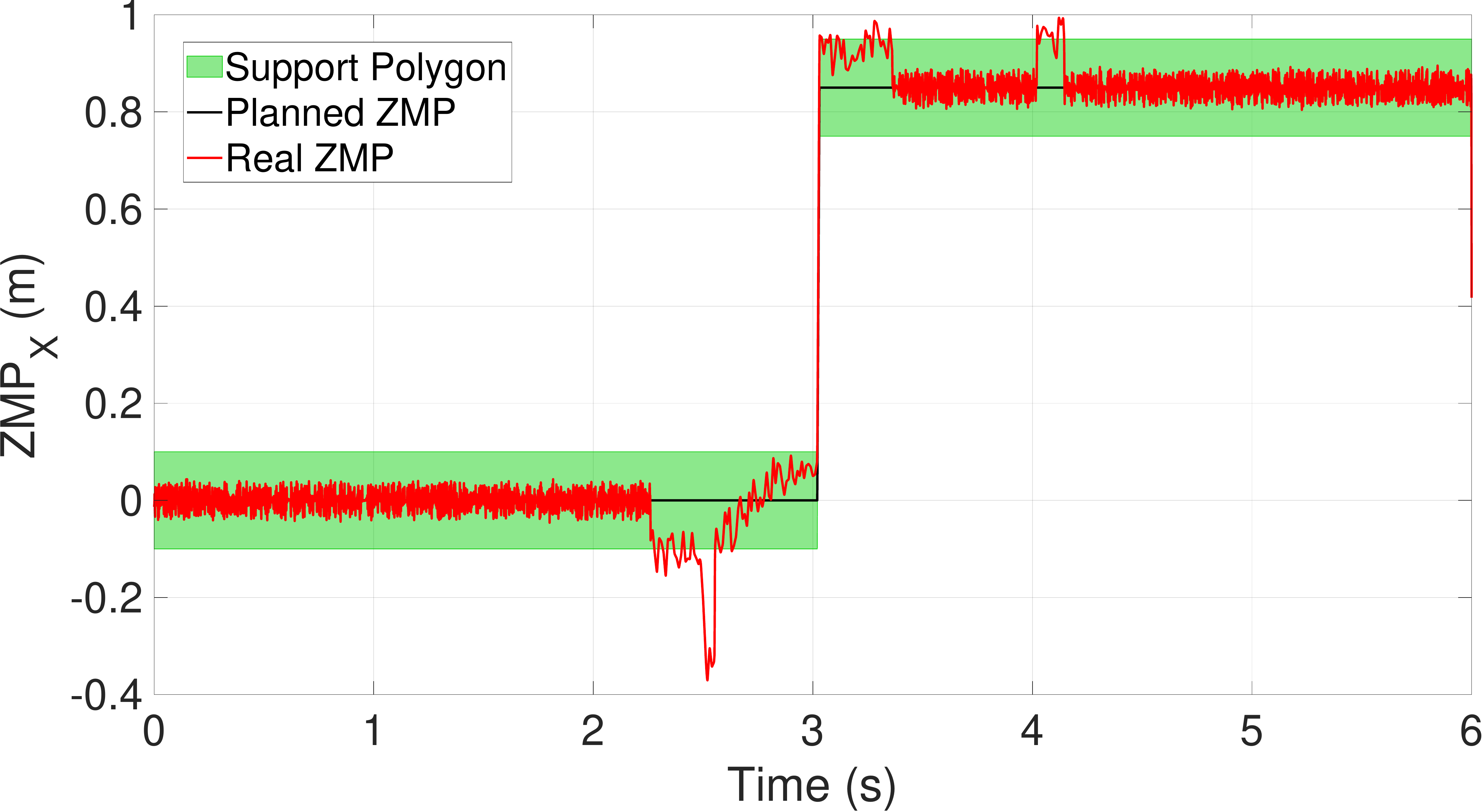}&
		\includegraphics[width=0.45\columnwidth, trim= 0cm 0cm 0cm 0cm,clip]{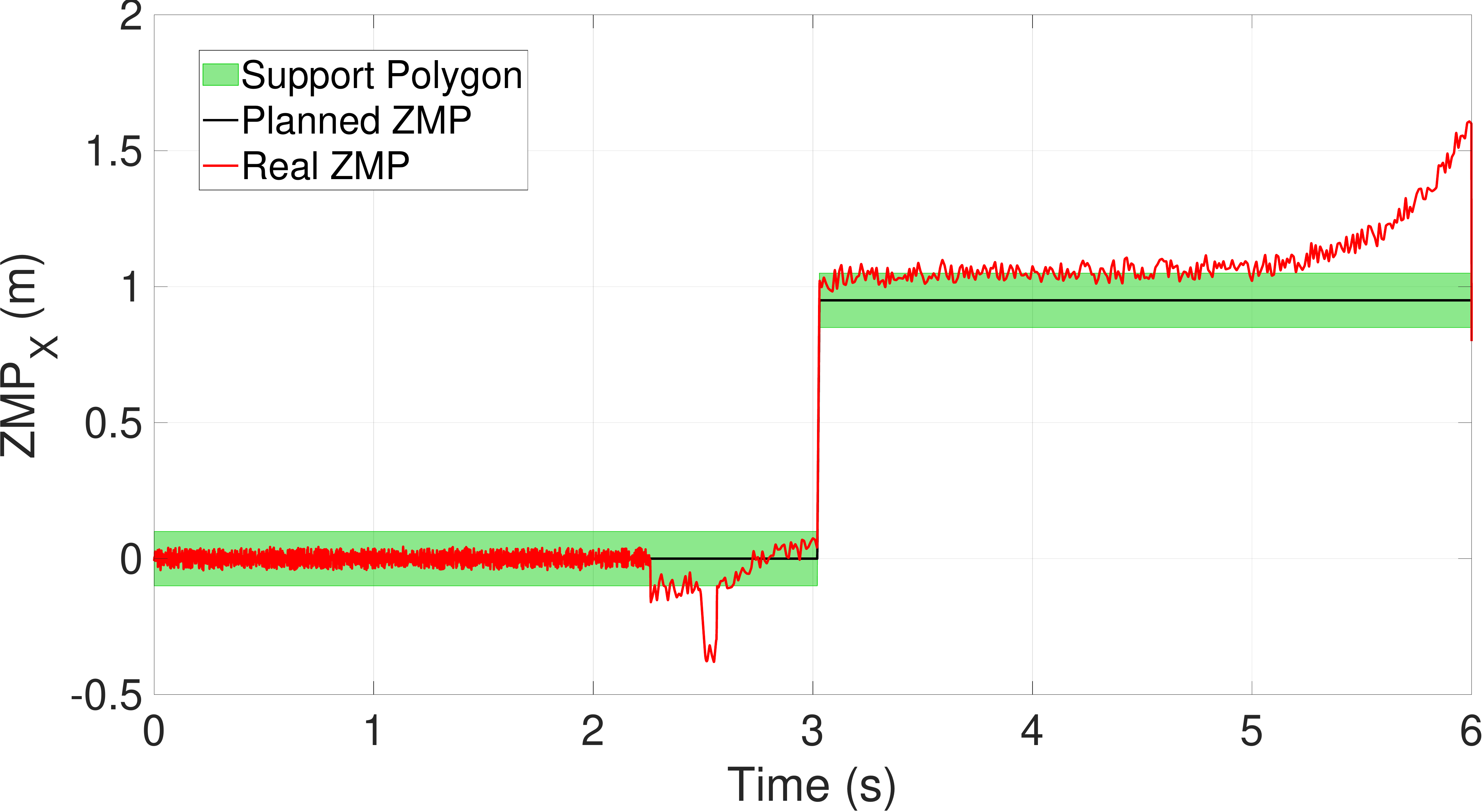}\\
		
		\includegraphics[width=0.45\columnwidth, trim= 0cm 0cm 0cm 0cm,clip]{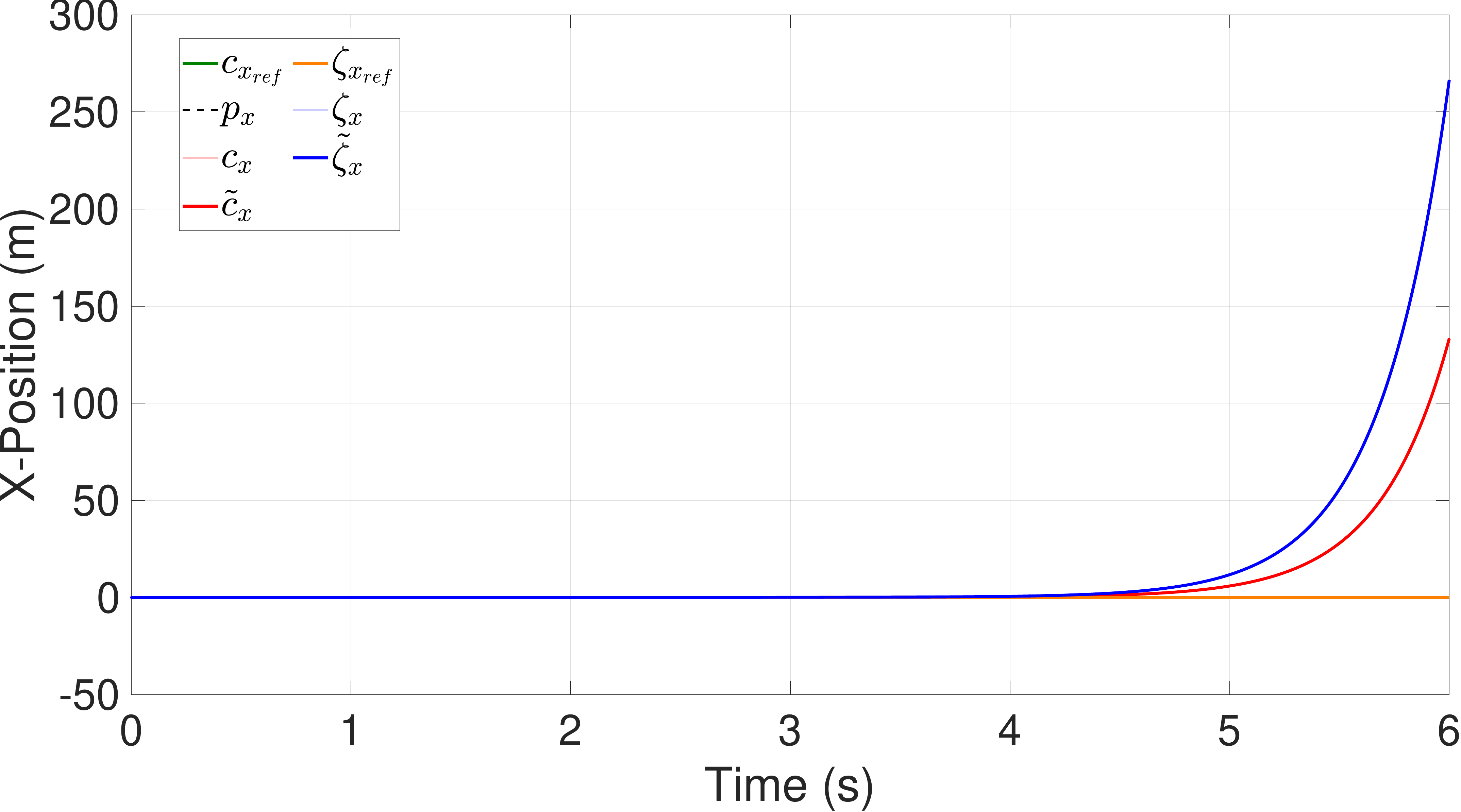}&
		\includegraphics[width=0.45\columnwidth, trim= 0cm 0cm 0cm 0cm,clip]{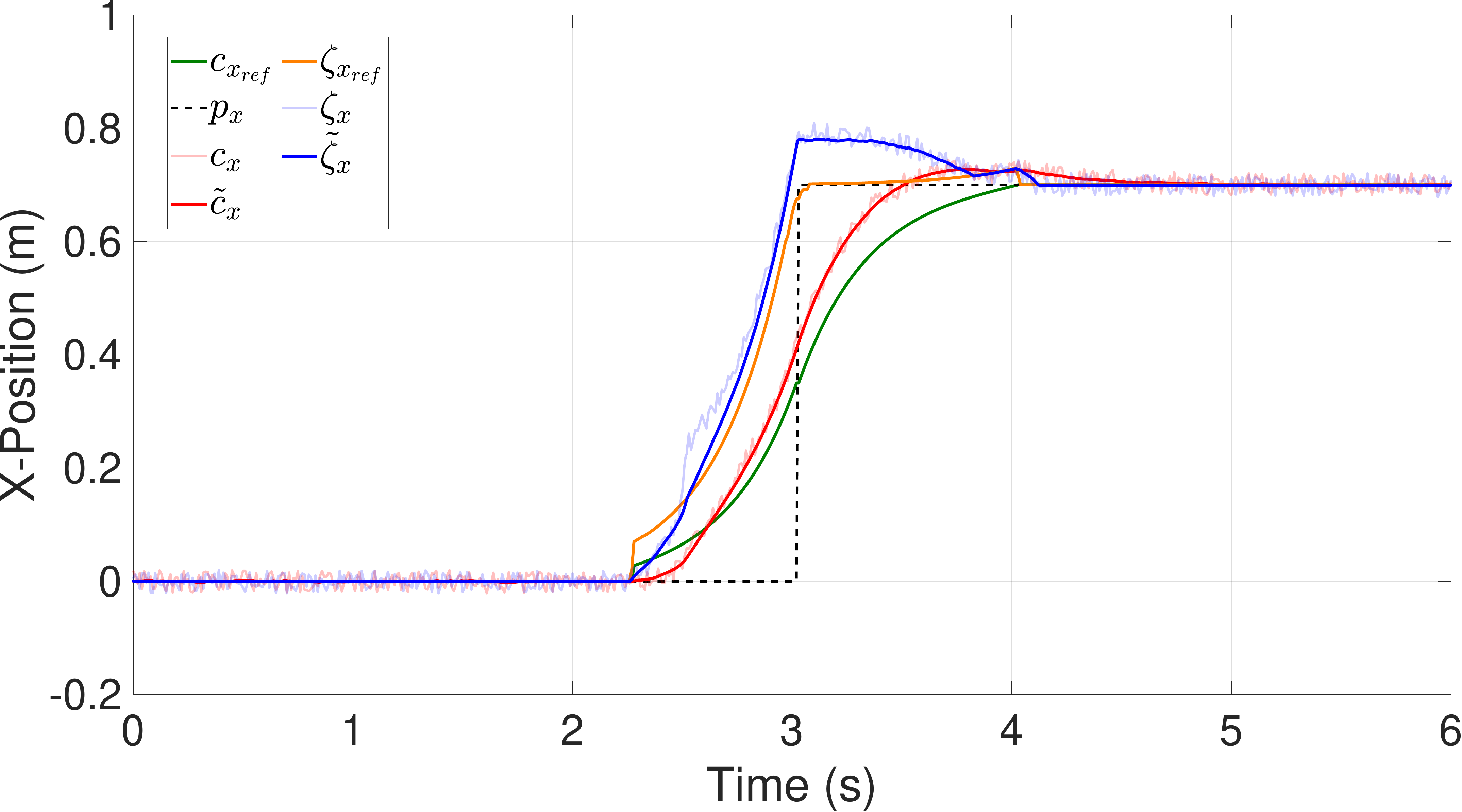}&
		\includegraphics[width=0.45\columnwidth, trim= 0cm 0cm 0cm 0cm,clip]{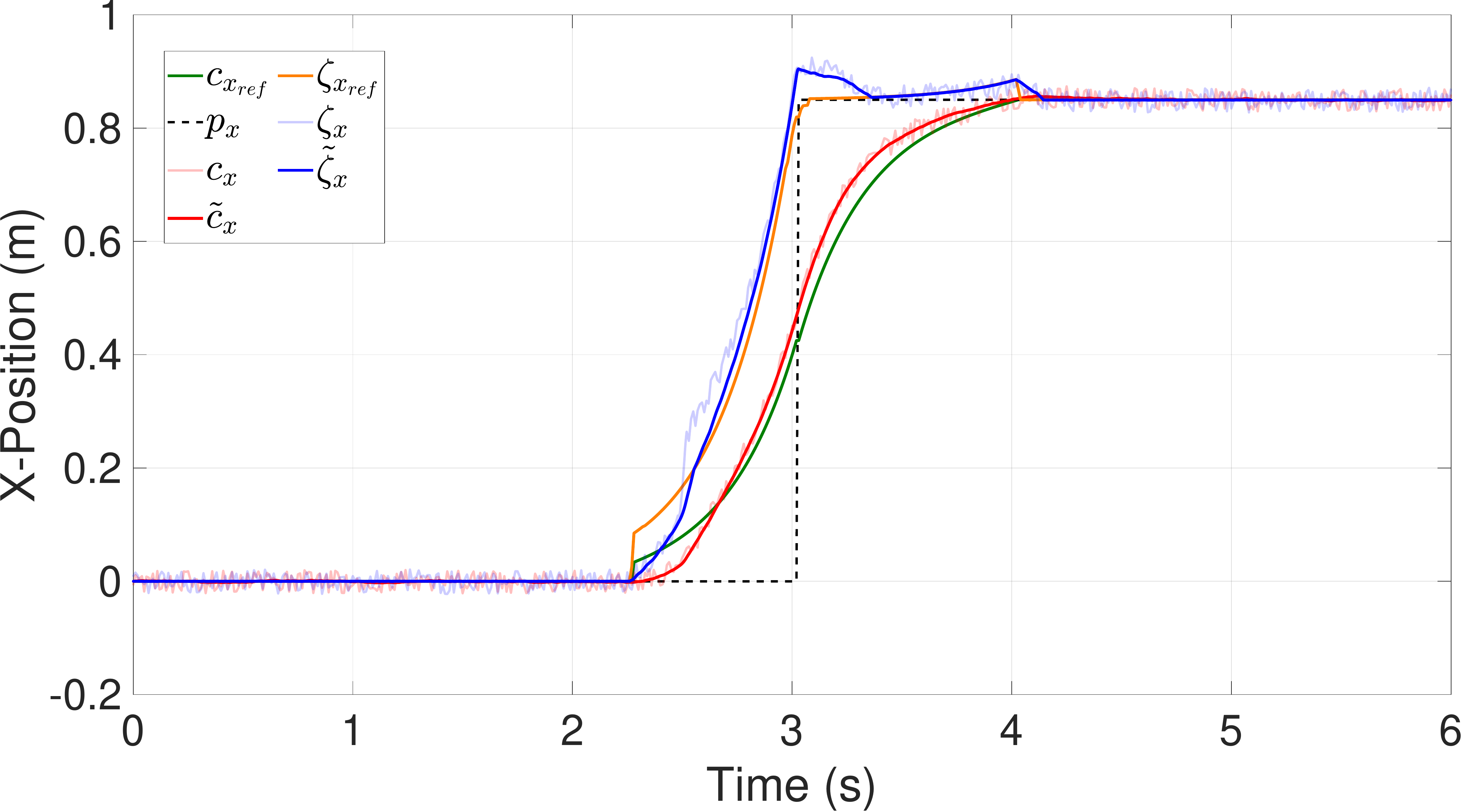}&
		\includegraphics[width=0.45\columnwidth, trim= 0cm 0cm 0cm 0cm,clip]{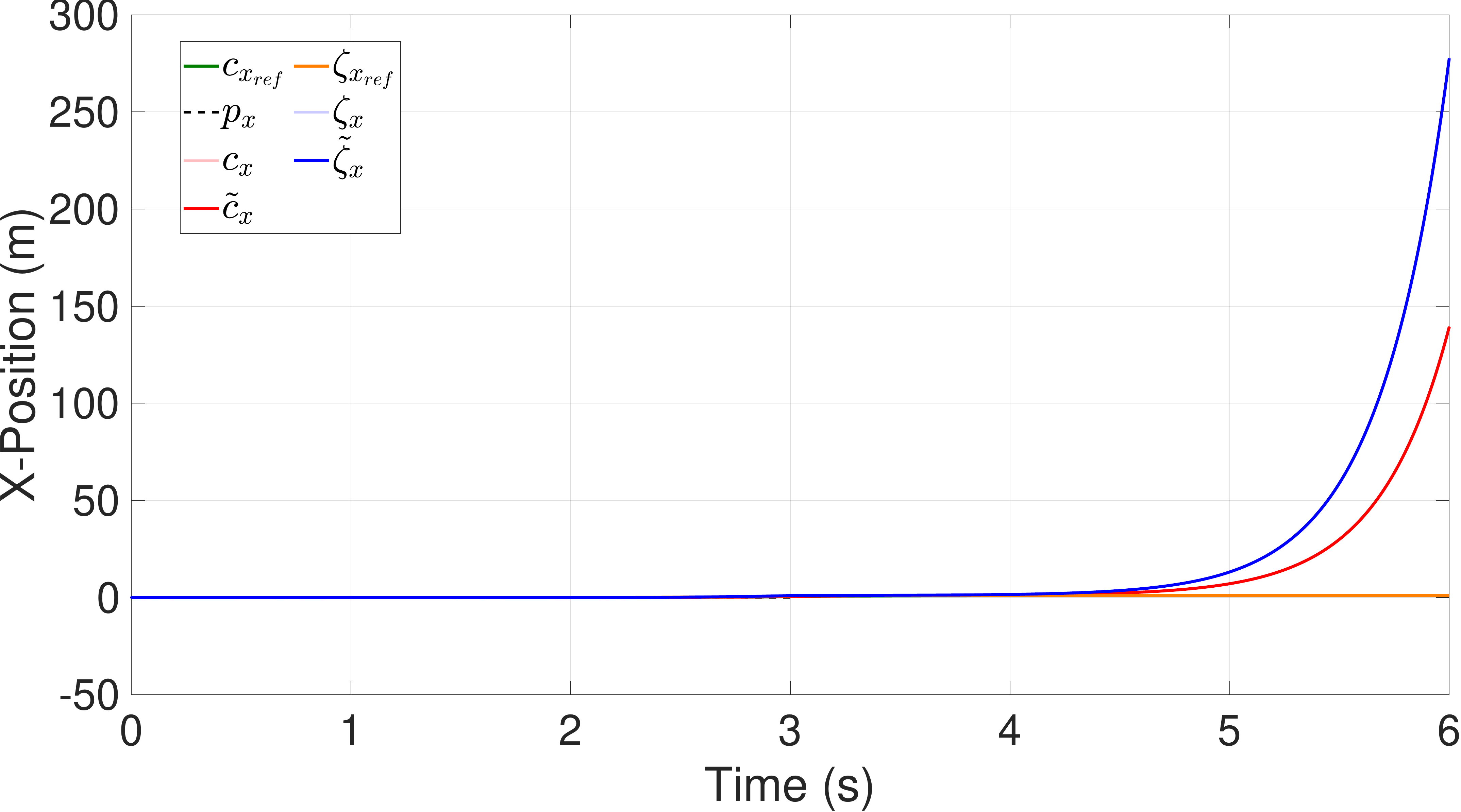}\\
		\small{$F_x = 326 N $}&\small{$F_x = 350 N$}&\small{$F_x = 400 N $}&\small{$F_x = 412 N $} \\
	\end{tabular}	
	\caption{ The simulation results of examining the robustness w.r.t. external disturbances. After applying a disturbance, the proposed planner modifies the landing location of the swing leg and change the reference trajectories to regain the stability of the robot.}		
	\label{fig:robust_ext}
\end{figure*}

\begin{figure*}[!t]
	\centering
	\begin{tabular}	{c c c c}			
		\includegraphics[width=0.45\columnwidth, trim= 0cm 0cm 0cm 0cm,clip]{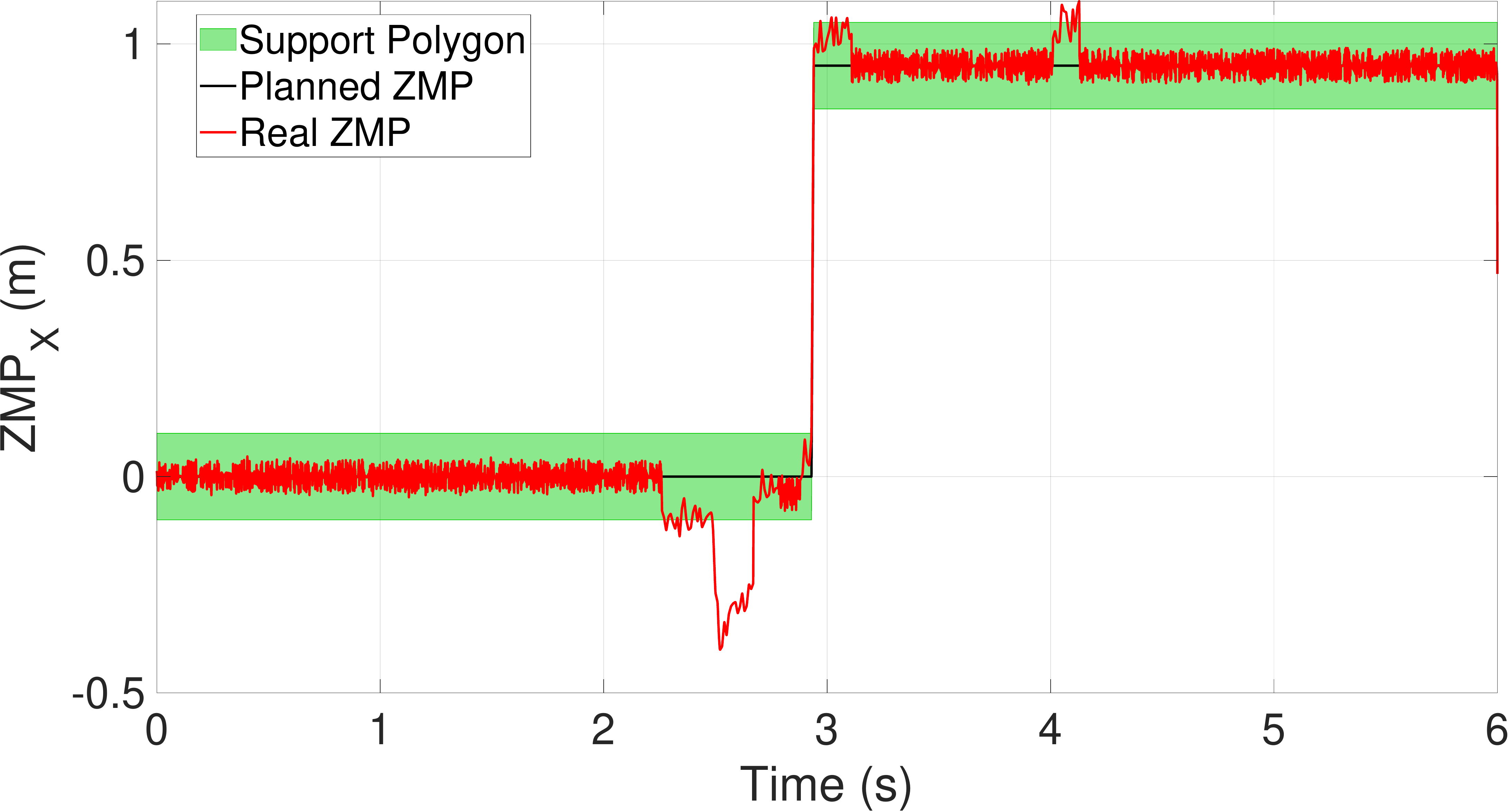}&
		\includegraphics[width=0.45\columnwidth, trim= 0cm 0cm 0cm 0cm,clip]{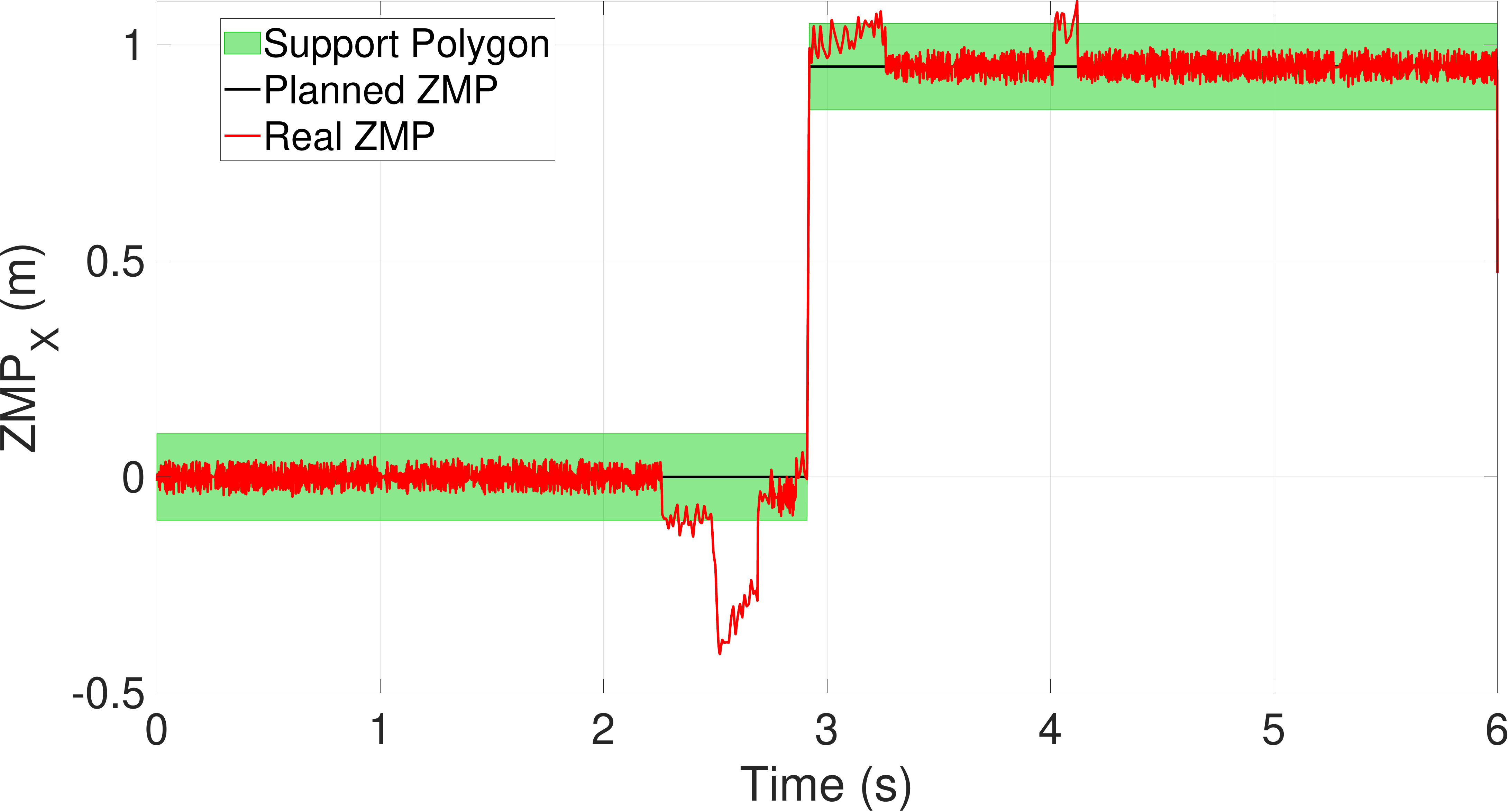}&
		\includegraphics[width=0.45\columnwidth, trim= 0cm 0cm 0cm 0cm,clip]{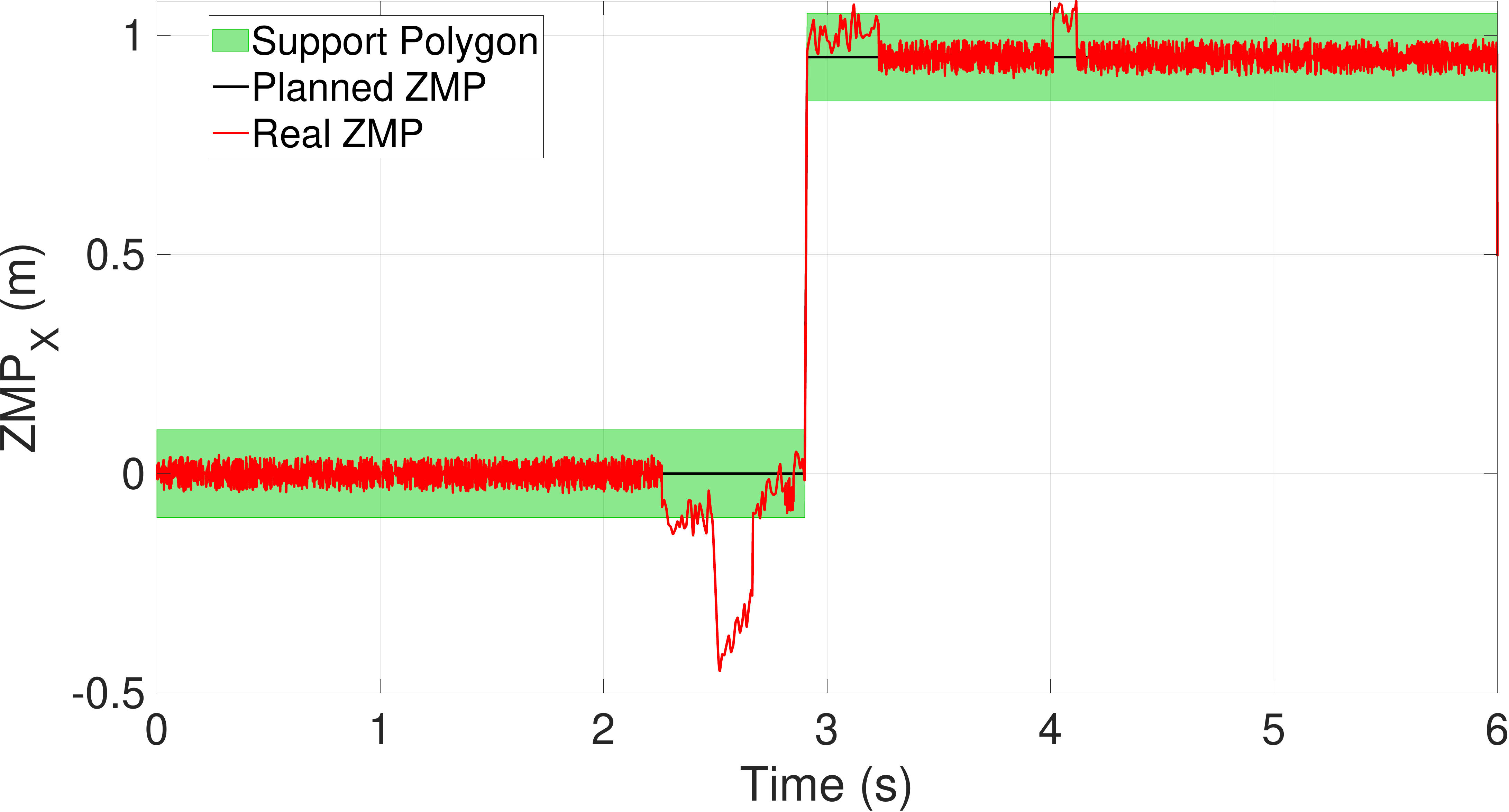}&
		\includegraphics[width=0.45\columnwidth, trim= 0cm 0cm 0cm 0cm,clip]{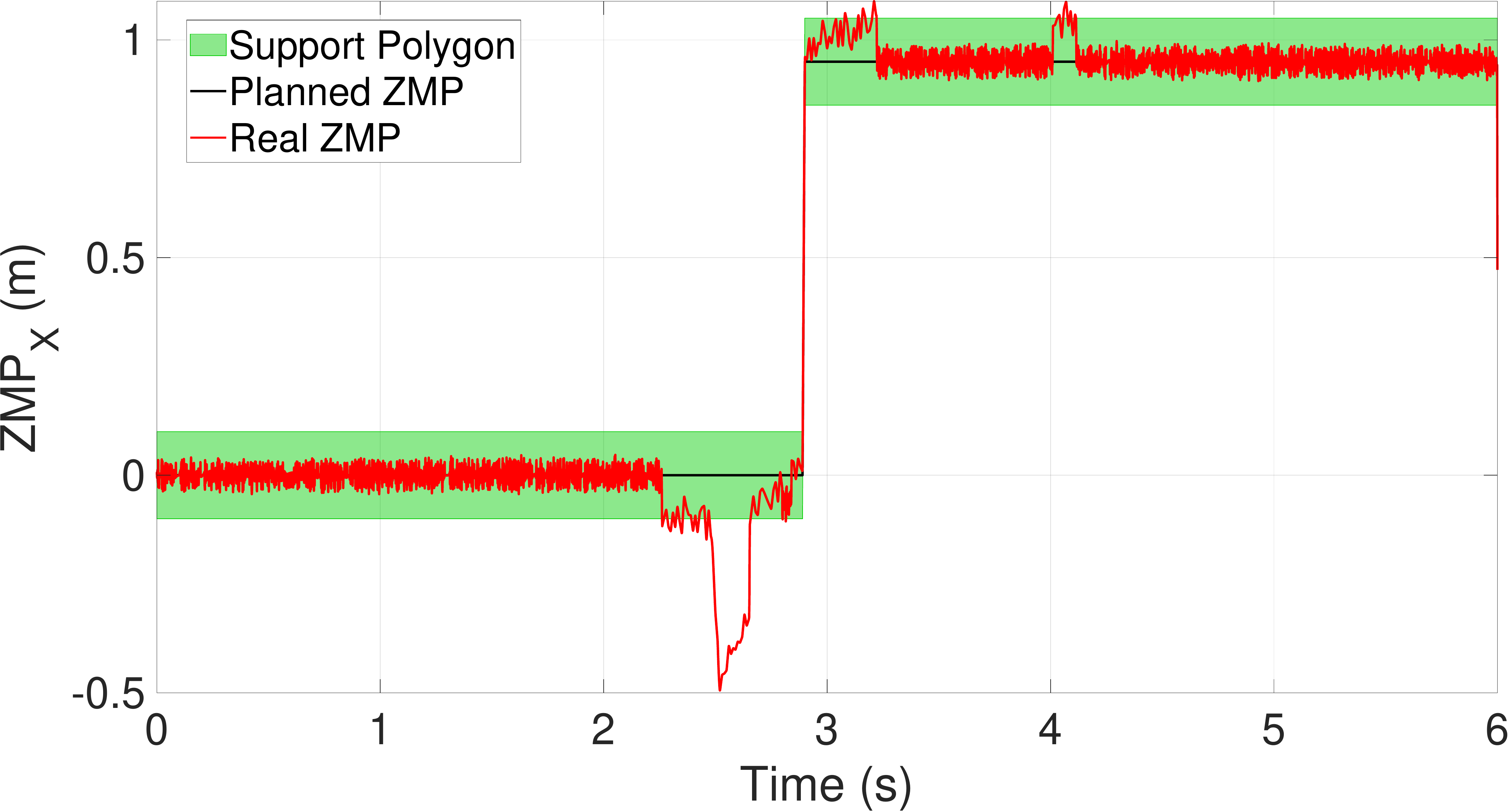}\\
		
		\includegraphics[width=0.45\columnwidth, trim= 0cm 0cm 0cm 0cm,clip]{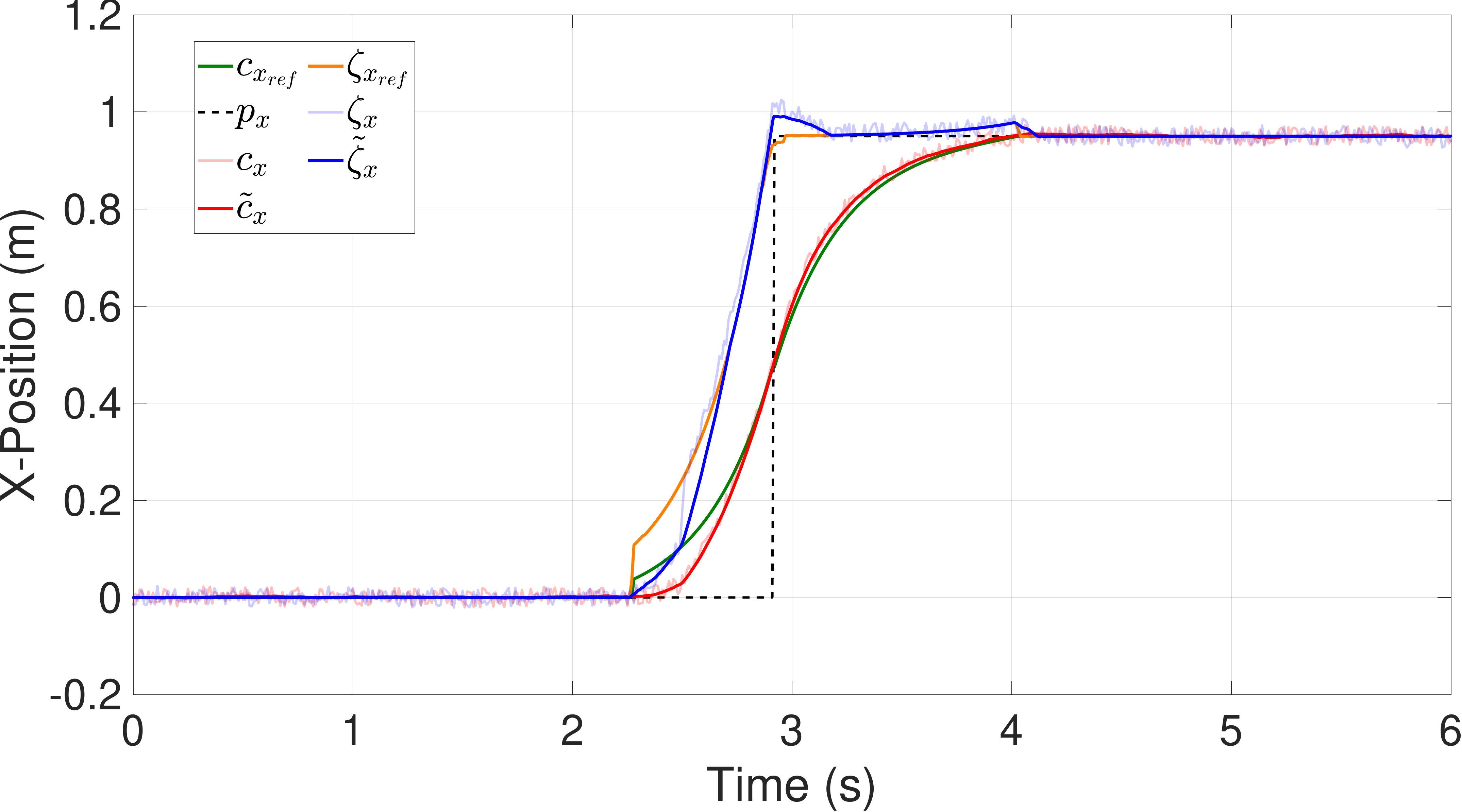}&
		\includegraphics[width=0.45\columnwidth, trim= 0cm 0cm 0cm 0cm,clip]{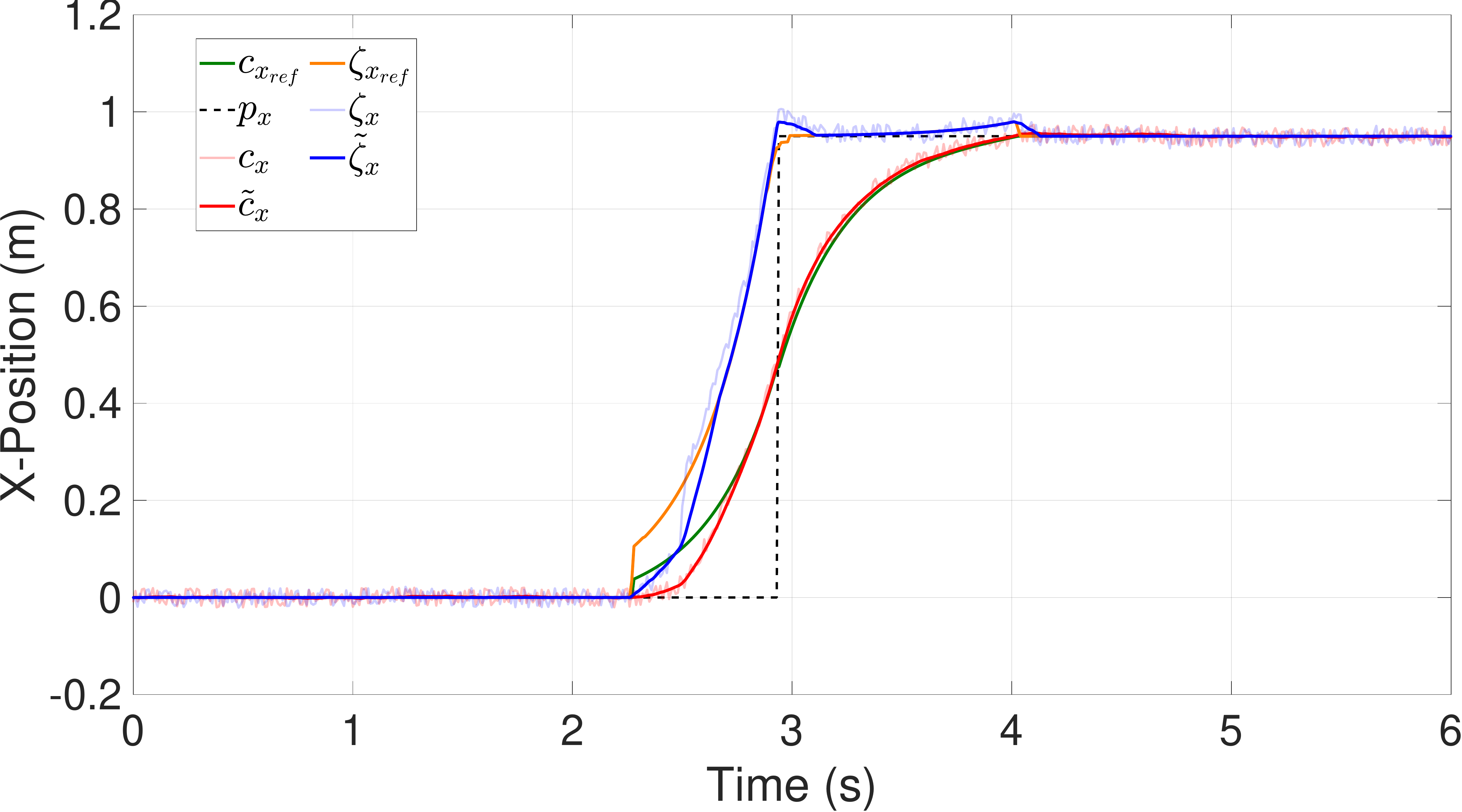}&
		\includegraphics[width=0.45\columnwidth, trim= 0cm 0cm 0cm 0cm,clip]{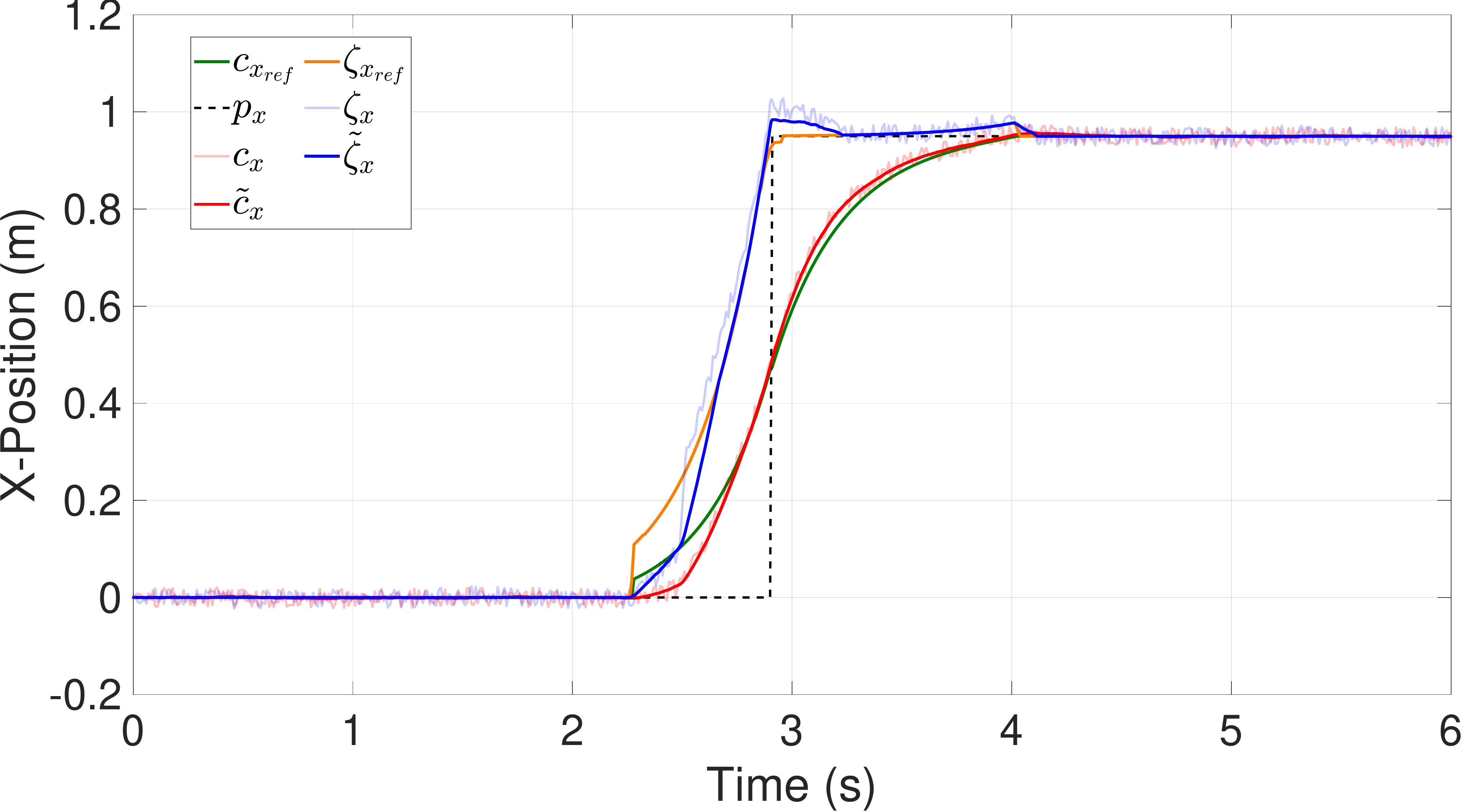}&
		\includegraphics[width=0.45\columnwidth, trim= 0cm 0cm 0cm 0cm,clip]{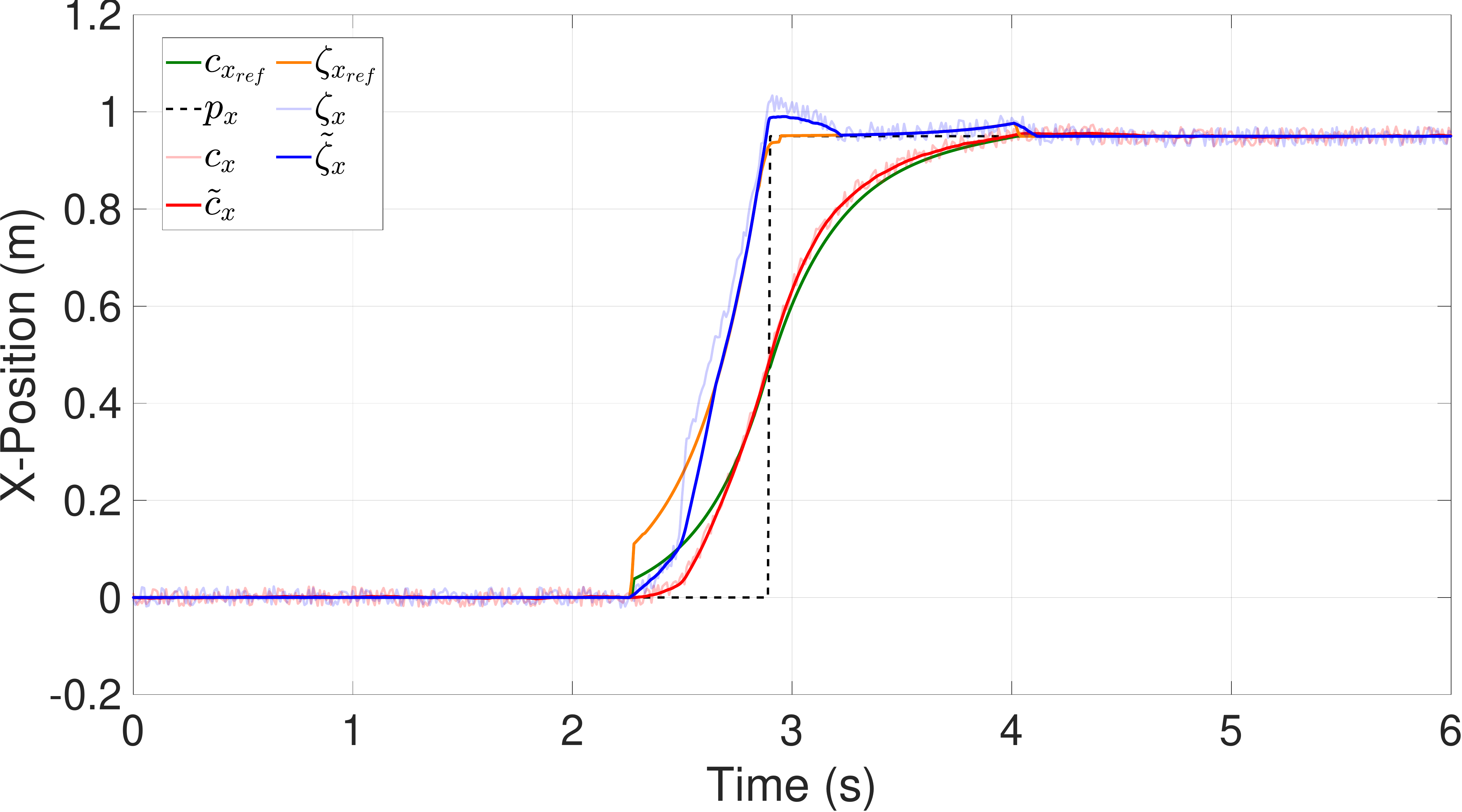}\\
		
		\small{$F_x = 412 N $}&\small{$F_x = 460 N$}&\small{$F_x = 515 N $}&\small{$F_x = 530 N $} \\
		
	\end{tabular}	
	\caption{ The simulation results of examining the robustness w.r.t. external disturbances. After applying a disturbance, the proposed planner modifies the time and the landing location of the swing leg and change the reference trajectories to regain the stability of the robot.}		
	\vspace{-5mm}	
	\label{fig:robust_ext2}
\end{figure*}

This scenario is focused on verifying the performance of the \textit{Next Step Adjusters} module. In this simulation, while the simulated robot is walking in place, a severe push with impact duration $10ms$ is applied to the robot's COM at~\mbox{t = $2.2s$}. The simulation is started by performing some trial and error to find the maximum amplitude of impact that the robot could keep the stability just by applying compensating torques. The simulation results showed that at $F_x=325N$ robot could not regain its stability and fall down~(see the first column of Fig.~\ref{fig:robust_ext}). After determining this value, firstly, the step time adjuster has been disabled, and the controller should keep the stability just by adjusting the next step location. This simulation has been repeated three times with different amplitudes of the impacts to find the maximum level of withstanding. It should be noted that during the simulations, the reference and measured ZMP, COM, DCM and the output of the controller have been recorded to analyze the behavior of the controller. The simulation results of this scenario are depicted in Fig.~\ref{fig:robust_ext}. The plots in each column represent all the actual and the reference trajectories. As is shown in the second column of this figure, after applying $F_x=350N$, the proposed controller changes the landing location of the swing leg to $0.71m$ to keep the stability. The simulation has been repeated with more severe impact~($F_x=400N$) and the results are depicted in the third column. As the results showed, the controller modified the landing location of the swing leg to $0.92m$ and could regain stability. According to the kinematic limitation of the simulated robot, the maximum step length that the simulated robot can take is $0.95m$. Hence to find the maximum impact that can be handled by step adjustment, the amplitude of the impact has been increased while the robot falls down. The results showed that $F_x=411N$ is the maximum level of impact that the simulated robot could handle using the proposed step adjustment strategy. According to the simulation results, the next step adjustment strategy improves the withstanding level of the robot up to 26\%. After evaluating the capability of the next step adjustment strategy, the step time adjustment has been enabled and the simulation has been repeated four times with more severe impacts. In these simulations, the output of the step time adjuster is considered to be saturated at $\pm 0.2s$ which means the step time adjuster can increase or decrease the step time up to $0.2s$. The simulation results are shown in Fig.~\ref{fig:robust_ext2}. The last simulation has been repeated~($F_x=412N$) to check the effectiveness of the step time adjustment. The simulation results are shown in the plots of the first column. The results showed that the step time adjuster decreased the step time by $0.12s$ and could keep the stability. The amplitude of the impact has been increased and the simulation has been repeated~($F_x=460N$). According to the simulation results, which are depicted in the plots of the second column, the simulated robot could regain its stability by decreasing $0.15s$ the step time. The amplitude of the impact has been increased and the simulation has been repeated until the output of the time step adjuster was not saturated. The simulation results of $F_x=515N$ and $F_x=530N$ are depicted in third and fourth columns, respectively. The simulation results showed that at $F_x=532N$, the output of the time step adjuster is saturated and robot could not keep its stability. The simulation results showed that adjusting step time improves the withstanding level of the simulated robot up to 29\%. According to the simulation results, the \textit{Next Step Adjuster} module improved the overall withstanding level of the is improved up to 63\%.

\section{DISCUSSIONS}
\label{sec:DISCUSSIONS}
Most of the presented works in the literature are based on online optimization methods~(e.g., QP, MPC) and due to iterative nature of these algorithms, their performance is sensitive to the computation power of the resources. In comparison with those works, the most important property of the proposed system is removing the online optimization of MPC and QP without reducing the adaptiveness level of the controller. Based on the presented simulation results in previous sections, the proposed system is capable of generating walking which is not only adaptive but also robust against uncertainties and disturbances. Besides, the proposed system is computationally supper fast; hence, it does not require specific resources. Unlike~\cite{morisawa2014biped,englsberger2015three} which are based on PID control, the proposed system is designed based on optimal control approach which is more effective, moreover, the proposed controller uses the step time adjustment strategy to improve the stability of the robot.

\section {Conclusion}
\label{sec:CONCLUSION}
In this paper, we have designed and developed a biped locomotion framework which was composed of four main modules that are organized in a hierarchical structure to fade the complexity of the walking problem and increase the flexibility of the framework. We explained the functionality of each module and illustrated how these modules interact with each other to generate stable locomotion. The fundamental modules of this framework are the \textit{low-level controller} and \textit{next step adjustment} which are responsible for motion tracking and keeping the stability. Particularly, the LIPM and DCM concepts were used to represent the overall dynamics of the system as a linear state-space system and using that, we formulated the walking problem as an LQG optimal controller to generate a robust control solution based on an offline optimization. The next step adjustment module was combined with the low-level controller module to improve the withstanding level of the framework by online modification of step time and location according to the measured DCM in each control cycle. The functionality of each module has been tested independently and the overall performance and robustness of the proposed framework were validated using performing several simulations. The simulation results showed that the next step adjustment module improves the overall withstanding level of the framework up to 63\%. 

Our future work includes examining the performance of the proposed framework while generating walking on uneven terrain. Additionally, we would like to port our framework to several real humanoid platforms to show the portability and flexibility of the proposed framework.

\section*{Acknowledgment}
This research is supported by Portuguese National Funds through Foundation for Science and Technology (FCT) through FCT scholarship SFRH/BD/118438/2016.

\bibliographystyle{IEEEtran}
\bibliography{IROS2019.bbl}

\end{document}